%% file: main.tex
\PassOptionsToPackage{table}{xcolor}
\documentclass[]{genesis}
\microtypesetup{expansion=false} % TinyTeX/pdfTeX cannot expand the bundled font.
\usepackage[page,header]{appendix}
\usepackage{booktabs}
\usepackage{multirow}
\usepackage{graphicx}
\usepackage{tabularx}
\usepackage{hyperref}
\usepackage{amsmath}
\usepackage{amssymb}
\usepackage{mathtools}
\usepackage{makecell}
\usepackage{pifont}
\usepackage{float}
\usepackage{enumitem}
\usepackage{inconsolata}
\usepackage[most]{tcolorbox}

\definecolor{genPurple}{HTML}{7C3AED}
\definecolor{genPurpleDark}{HTML}{5B21B6}
\definecolor{ourscolor}{HTML}{EFE9FD}

\definecolor{promptbg}{HTML}{F8F5FC}
\definecolor{promptbd}{HTML}{7C3AED}
\newtcolorbox{promptbox}[1][]{%
  enhanced, breakable,
  colback=promptbg, colframe=promptbd,
  boxrule=0.4pt, arc=2pt, left=4pt, right=4pt, top=3pt, bottom=3pt,
  fonttitle=\bfseries\small, coltitle=white,
  colbacktitle=promptbd,
  attach boxed title to top left={xshift=4pt, yshift=-2pt},
  boxed title style={size=small, colframe=promptbd, sharp corners},
  #1
}

\setlist[itemize]{leftmargin=15pt}
\RequirePackage{xspace}
\makeatletter
\DeclareRobustCommand\onedot{\futurelet\@let@token\@onedot}
\def\@onedot{\ifx\@let@token.\else.\null\fi\xspace}

\makeatother



\title{{\fontsize{14.5}{18}\selectfont \textcolor{genPurple}{Genesis}: Visual Augmentation Primitives for Multimodal Chain-of-Thought Reasoning}\\[5pt] {\fontsize{11}{13}\selectfont\sffamily\itshape\color{genPurple}Accepted to the EMNLP 2026 Main Conference}}

\author{\textbf{Fan Yang}\textsuperscript{\rm 1,3,*}\hspace{0.6em}
\textbf{Lifeng Chen}\textsuperscript{\rm 4,*}\hspace{0.6em}
\textbf{Yipu Wang}\textsuperscript{\rm 2,3,*}\hspace{0.6em}
\textbf{Han Wang}\textsuperscript{\rm 3,*}\hspace{0.6em}
\textbf{Feilong Chen}\textsuperscript{\rm 7,$\diamondsuit$,$\dagger$}\hspace{0.6em}
\textbf{Yidong Ma}\textsuperscript{\rm 7}\\
\textbf{Zhiyang Chen}\textsuperscript{\rm 5}\hspace{0.6em}
\textbf{Chaoyang Zhao}\textsuperscript{\rm 1,$\dagger$}\hspace{0.6em}
\textbf{Lingxiang Wu}\textsuperscript{\rm 1}\hspace{0.6em}
\textbf{Hanqi Jiang}\textsuperscript{\rm 4}\hspace{0.6em}
\textbf{Yibo Chen}\textsuperscript{\rm 7}\hspace{0.6em}
\textbf{Xu Tang}\textsuperscript{\rm 7}\\
\textbf{Yao Hu}\textsuperscript{\rm 7}\hspace{0.6em}
\textbf{Ming Tang}\textsuperscript{\rm 1}\hspace{0.6em}
\textbf{Xiaojie Jin}\textsuperscript{\rm 4}\hspace{0.6em}
\textbf{Xiaolong Zheng}\textsuperscript{\rm 2,3}\hspace{0.6em}
\textbf{Jinqiao Wang}\textsuperscript{\rm 1,3,6,$\dagger$}\\[2mm]
{\small
\textsuperscript{\rm 1}Foundation Model Research Center, Institute of Automation, Chinese Academy of Sciences\\
\textsuperscript{\rm 2}State Key Laboratory of Multimodal Artificial Intelligence Systems,\\
Institute of Automation, Chinese Academy of Sciences\quad
\textsuperscript{\rm 3}University of Chinese Academy of Sciences\\
\textsuperscript{\rm 4}Beijing Jiaotong University\quad
\textsuperscript{\rm 5}Independent Researcher\quad
\textsuperscript{\rm 6}Wuhan AI Research\quad
\textsuperscript{\rm 7}Xiaohongshu Inc.}\\[3mm]
\includegraphics[width=\linewidth,height=28mm,keepaspectratio]{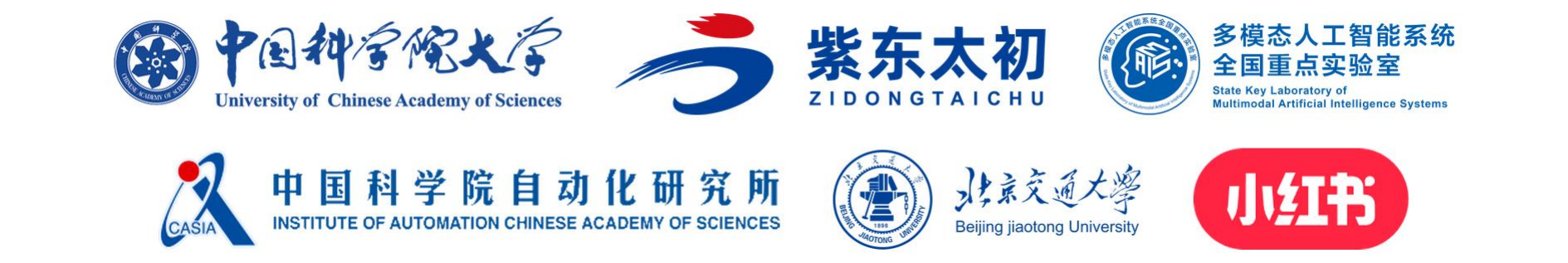}\\[1mm]
{\footnotesize \textsuperscript{*}Equal contribution\quad \textsuperscript{$\diamondsuit$}Project leader\quad \textsuperscript{$\dagger$}Corresponding author\\
\textsuperscript{*}Work done while the four co-first authors were interns at Xiaohongshu.}}

\renewcommand{\abstractinfont}{\fontsize{9}{10.5}\selectfont}
\abstract{
Recent multimodal large language models achieve promising results on visual reasoning tasks, approaching human-level performance on existing STEM benchmarks. However, these capabilities degrade under complex real-world scenes. We identify that chain-of-thought reasoning remains largely confined to the linguistic space, rarely anchoring each step in visual evidence. The main cause is the absence of such data during training. In this work, we introduce a training data augmentation pipeline, \textbf{Visual Augmentation Primitives} (VAP), which leverages state-of-the-art vision generative models to synthesize diverse photorealistic training samples and corresponding reasoning text. VAP introduces some basic operators and applies them to a seed image and gets a variant. It also constructs coherent multimodal reasoning chains throughout the generation process. Such training data force models to perceive visual content and perform logical reasoning according to them at every step. We further introduce multimodal chain-of-thought rubric-based reinforcement learning to enhance this. It rewards not only text reasoning but also visual reasoning by scoring intermediate reasoning steps for logical validity, completeness, and consistency. Moreover, we construct the UNREAL benchmark to systematically evaluate visual reasoning on hyperrealistic scenes across controllable difficulty levels. Experiments show that a 4B model built on this framework surpasses 8B baselines, with performance improving consistently as the model scales to 32B, validating the scaling law for visual reasoning. All models, data, and code will be publicly released.

}

\begin{document}
\maketitle

\clearpage
% Keep the table of contents scannable: list sections and subsections only.
\setcounter{tocdepth}{2}
\tableofcontents
\setcounter{tocdepth}{2}
\clearpage
\input{chapters/1_introduction}
\input{chapters/2_related_work}

\input{chapters/3_vap}

\input{chapters/4_rl}
\input{chapters/5_unreal}
\input{chapters/6_experiments}

\input{chapters/7_conclusion}
\input{chapters/acknowledgments}
\FloatBarrier
\bibliographystyle{plainnat}
\bibliography{references}
\clearpage
\input{chapters/appendix}

\end{document}

%% file: chapters/1_introduction.tex
\section{Introduction}
\label{sec:intro}

% Recent multimodal large language models (MLLMs) achieve strong performance on visual reasoning tasks such as geometry, function analysis, and quantitative problem solving. Representative systems such as GPT-5.5~\cite{openai2025gpt55}, Gemini 3.5~\cite{google2025gemini35}, Claude 4.7~\cite{anthropic2025claude47}, and Qwen 3.7~\cite{qwen2025qwen37} establish impressive results on complex problems in mathematics, physics, and chemistry that combine textual descriptions with visual content such as diagrams and plots~\cite{lu2024mathvista, yue2024mmmu, lu2022scienceqa}. The reasoning chains they produce now display coherent step-by-step deduction together with meaningful self-reflection, marking a clear emergent capability~\cite{openai2024o1, deepseek2025r1}.

Recent multimodal large language models (MLLMs) achieve strong performance on visual reasoning tasks such as geometry, function analysis, and quantitative problem solving. Representative models such as GPT-5.5~\cite{openai2025gpt55}, Gemini 3.5~\cite{google2025gemini35}, Claude 4.7~\cite{anthropic2025claude47}, and Qwen 3.7~\cite{qwen2025qwen37} establish impressive results on complex problems in mathematics, physics, and chemistry that combine textual descriptions and visual evidence~\cite{yang2026seg, lu2024mathvista, yue2024mmmu, lu2022scienceqa}. As illustrated in Figure~\ref{fig:first}(A), these models already achieve strong generalization across a range of visual reasoning tasks.

% Recent multimodal large language models (MLLMs) achieve strong performance on STEM-style reasoning tasks such as geometry, function analysis, and quantitative problem solving. Representative systems such as GPT-5.5~\cite{openai2025gpt55}, Gemini 3.5~\cite{google2025gemini35}, Claude 4.7~\cite{anthropic2025claude47}, and Qwen 3.7~\cite{qwen2025qwen37} establish impressive results on complex problems in mathematics, physics, and chemistry that combine textual descriptions with visual content such as diagrams and plots~\cite{lu2024mathvista, yue2024mmmu, lu2022scienceqa}. The reasoning chains they produce now display coherent step-by-step deduction together with meaningful self-reflection, marking a clear emergent capability~\cite{openai2024o1, deepseek2025r1}.

% 近期的多模态大语言模型（MLLMs）在几何、函数分析、定量问题求解等 STEM 类推理任务上取得了优异的性能。GPT-5.5、Gemini 3.5、Claude 4.7、Qwen 3.7 等代表性系统在结合文本描述与图表、函数图等视觉内容的复杂数学、物理与化学问题上取得了令人瞩目的成绩。它们所产出的推理链如今呈现出连贯的逐步推导与有意义的自我反思，标志着一种清晰的涌现能力。

\begin{center}
\captionsetup{hypcap=false}
\includegraphics[width=0.98\linewidth]{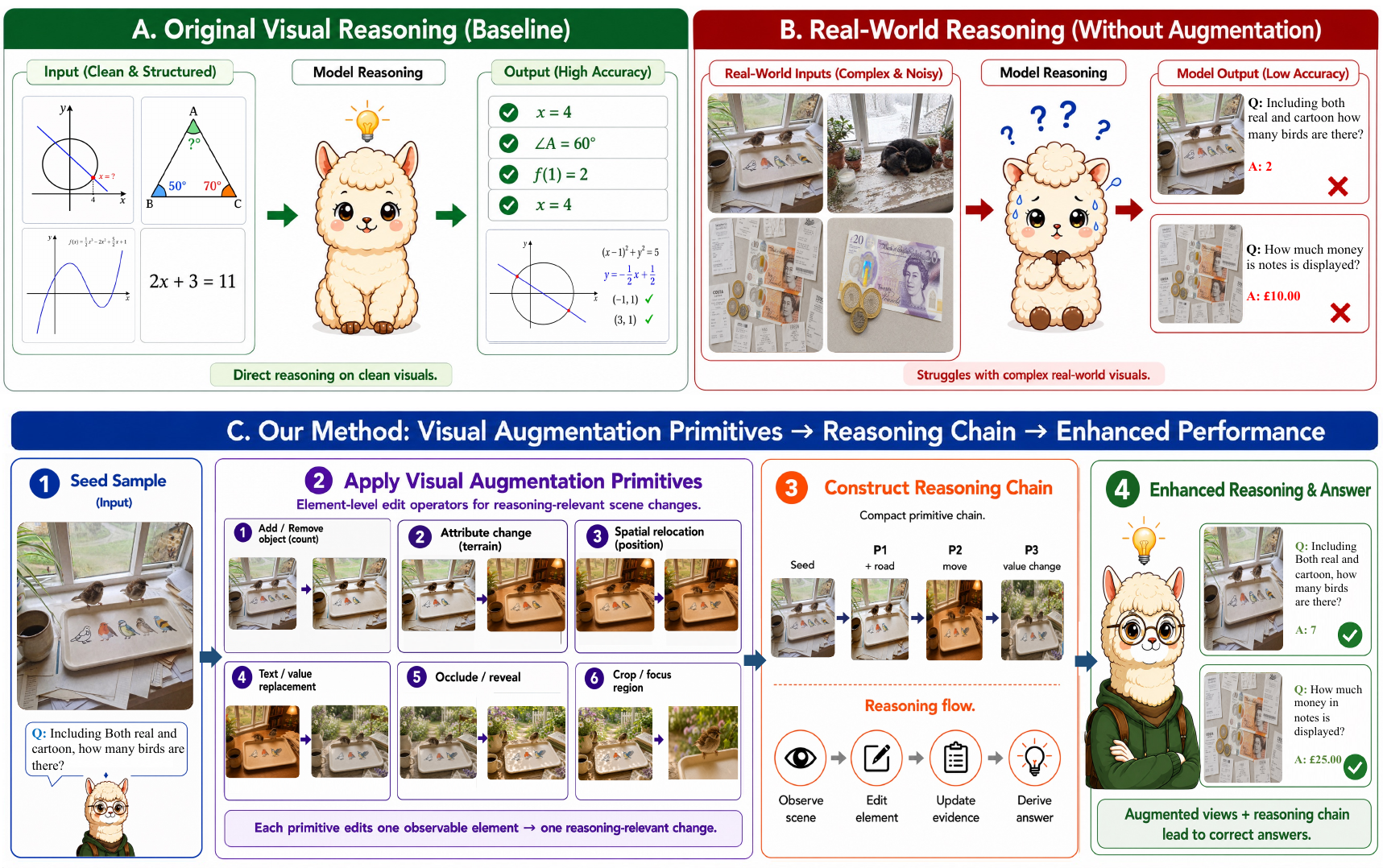}
\captionof{figure}{Motivation for \textbf{Visual Augmentation Primitives}. \textbf{(A)} Existing MLLMs already achieve strong generalization on a range of visual reasoning tasks with structured scenes. \textbf{(B)} These capabilities degrade on complex real-world scenes, where models produce incoherent reasoning chains that are rarely anchored in visual evidence. \textbf{(C)} VAP applies controllable scene-editing primitives to a seed image, so that every step from holistic scene understanding to localized edit regions stays consistent with its corresponding reasoning step.}
\label{fig:first}
\end{center}

However, as shown in Figure~\ref{fig:first}(B), these capabilities degrade on complex real-world scenes, where those models frequently produce incoherent reasoning chains and misread the spatial relations between objects~\cite{fu2024blink, liu2023visualspatial}. We attribute this degradation to two intertwined causes. First, chain-of-thought reasoning remains largely confined to the linguistic space, with intermediate steps rarely anchored in multimodal visual evidence, leading to hallucinated reasoning that compounds into incorrect final answers~\cite{tong2024eyeswideshut}. Second, the scarcity of multimodal chain-of-thought training data prevents MLLMs from leveraging scaling laws to improve multi-hop visual reasoning capabilities on real-world scenes. This raises a fundamental research question: \textbf{\textit{Can we extend the emergent reasoning capabilities of MLLMs to complex real-world scenes?}}

In this paper, we introduce \textbf{Visual Augmentation Primitives} (VAP), a set of controllable, generation-based scene editing primitives. VAP defines global scene analysis, key element identification, and edit instructions as atomic operations, each explicitly aligned with a corresponding step in the multimodal chain-of-thought. As illustrated in Figure~\ref{fig:first}(C), this design ensures that every step from holistic scene understanding to localized edit regions remains consistent with its corresponding reasoning step. Applying these primitives through instruction editing models across diverse visual contexts produces large-scale training data paired with multimodal chain-of-thought supervision, laying the foundation for scaling visual reasoning. Moreover, we further observe that existing RLVR optimizes only final-answer correctness and lacks supervision over intermediate reasoning, which easily leads to inconsistencies between the chain-of-thought and the final answer. We therefore propose \textbf{Genesis}, a multimodal reasoning language model that combines VAP with multimodal chain-of-thought rubric-based reinforcement learning. By extending rubric-based reward design~\cite{lightman2024lets, wang2024mathshepherd} from the text-only domain to visual reasoning, the rubric rewards can score each intermediate reasoning step along three dimensions, namely logical consistency, reasoning completeness, and multi-step process coherence.

Meanwhile, to test whether models truly rely on visual evidence for reasoning, we construct the \textbf{UNREAL} benchmark. UNREAL deliberately designs scenes that contradict common sense, so that models cannot arrive at correct answers by relying solely on language priors but must anchor every reasoning step in the visual content actually presented in the image. The benchmark generates image and question pairs with controllable difficulty levels, enabling systematic evaluation of both hallucination robustness~\cite{li2023pope, guan2024hallusionbench} and visual reasoning capability. Experiments validate the effectiveness of our approach. A 4B-parameter Genesis model surpasses 8B-parameter baselines on multiple tasks, and scaling to 32B achieves significant improvements across visual reasoning and VQA benchmarks.

% Leveraging the same generative capability, we introduce the \textbf{UNREAL} benchmark, specifically designed to stress-test MLLM reasoning on hyperrealistic scenes. Unlike manually curated benchmarks limited in both diversity and difficulty, UNREAL is constructed through the same instruction-guided image generation pipeline that powers Genesis training, producing image and question pairs with verifiable properties along controllable dimensions and scalable difficulty levels. Based on UNREAL, we evaluate model robustness against visual hallucinations~\cite{li2023pope, guan2024hallusionbench} together with the ability to capture real-world logical patterns. Experiments validate the effectiveness of our approach. A 4B-parameter Genesis model surpasses 8B-parameter baselines on multiple tasks, demonstrating exceptional data efficiency, and the same recipe scales to larger models with consistent gains across visual reasoning and VQA benchmarks. These results indicate that combining instruction-guided data synthesis with rubric-based reinforcement learning provides an effective, data-centric path toward scalable visual reasoning in complex real-world scenarios.

% 借助同样的生成能力，我们提出 \textbf{UNREAL} 基准，专门用于在超真实场景上对 MLLM 推理进行压力测试。与多样性和难度均受限的人工整理基准不同，UNREAL 通过与 Genesis 训练所用的同一条指令引导图像生成流水线构建，沿可控维度生成具备可验证属性的图像与问题对，并可在难度等级上灵活扩展。基于 UNREAL，我们评估模型对视觉幻觉的鲁棒性，并同时考察其对真实世界逻辑模式的理解能力。实验结果验证了所提方法的有效性。4B 参数的 Genesis 模型在多个任务上超越 8B 参数基线，展现出极高的数据效率。相同方法扩展到更大模型时也在视觉推理与视觉问答基准上取得稳定提升。这些结果表明，将指令引导的数据合成与基于评分标准的强化学习相结合，是面向复杂真实世界场景实现可扩展视觉推理的有效路径。

\begin{itemize}
  \item We introduce \textbf{Visual Augmentation Primitives}, which defines the visual elements of real-world images as editing primitives. Instruction-guided generative models apply these primitives to produce photorealistic image variants paired with multimodal reasoning chains, requiring models to unify visual perception with logical reasoning at every step.

    \item We introduce multimodal chain-of-thought rubric-based reinforcement learning, which extends rubric-based reward design from the text-only domain to visual reasoning, scoring intermediate reasoning steps for logical validity, completeness, and consistency.

  \item We construct the \textbf{UNREAL} benchmark to systematically evaluate visual reasoning on scenes that contradict text priors across diverse visual contexts. A 4B Genesis model surpasses 8B baselines, and scaling to 32B achieves significant improvements across visual reasoning and VQA benchmarks. All models, data, and code will be publicly released.
\end{itemize}

% 我们的主要贡献如下。
% \begin{itemize}
%     \item 我们将训练数据稀缺识别为复杂视觉推理的主要瓶颈，并通过一条基于指令引导的数据构造流水线加以解决，从极少量种子样本扩展训练数据，无需依赖人工标注或传统数据增强。
%     \item 我们提出 \textbf{Genesis}，将该流水线（包含逻辑答案扩展与场景方差生成）与基于评分标准的强化学习相结合，使推理监督从仅依赖最终答案推进到过程级监督。
%     \item 我们提出 \textbf{UNREAL} 基准，对超真实场景下的视觉推理进行压力测试，并展示出色的数据效率，4B 参数的 Genesis 模型即可超越 8B 基线，且该方法能稳定扩展到更大模型。
% \end{itemize}

%% file: chapters/2_related_work.tex
\section{Related Work}

\subsection{Visual Reasoning Benchmarks and Their Limitations}
A wide range of benchmarks evaluate visual reasoning in multimodal models. Broad assessments such as SEED-Bench~\cite{li2024seedbench} and MMBench~\cite{liu2024mmbench} cover general capabilities, while specialized suites including MathVista~\cite{lu2024mathvista}, MMMU~\cite{yue2024mmmu}, and ScienceQA~\cite{lu2022scienceqa} target STEM problem solving. Most rely on multiple-choice formats or templated questions and struggle to probe genuine visual understanding beyond their authored templates. Diagnostic suites such as BLINK~\cite{fu2024blink} and HallusionBench~\cite{guan2024hallusionbench} further reveal that frontier MLLMs still falter on basic perception and confidently hallucinate on real-world images~\cite{tong2024eyeswideshut}. Recent efforts push toward real-world dynamics and finer-grained visual grounding, including reasoning over physical visual transformations~\cite{ji2025visualtrans} and cross-view point correspondence~\cite{wang2025towards}, yet none offers scalable, controllable supervision that ties every reasoning step to observable scene evidence. Our UNREAL benchmark complements this line by stress-testing reasoning on hyperrealistic generated scenes with controllable difficulty, enabling systematic evaluation across scalable difficulty levels.

\subsection{Chain-of-Thought Supervision and Reasoning Consistency}
Chain-of-thought (CoT) prompting~\cite{wei2022cot} has become central to multi-step reasoning, with multimodal variants such as Multimodal-CoT~\cite{zhang2023mmcot} and Visual CoT~\cite{shao2024visualcot, yang2026tracevision} extending it to vision-language inputs. A persistent limitation is that reasoning chains in MLLMs often drift toward language priors rather than observable visual evidence, and this drift compounds as chains grow longer~\cite{tong2024eyeswideshut}. Process-level supervision approaches such as Let's Verify Step by Step~\cite{lightman2024lets} and Math-Shepherd~\cite{wang2024mathshepherd} address the same problem in the text domain by rewarding intermediate steps, and toolkits such as PRM-as-a-Judge~\cite{liu2026prm} begin to systematize process assessment for embodied tasks, but such process reward models have rarely been applied to visual reasoning at scale. Genesis brings this direction into the visual domain by pairing every synthetic scene with reasoning chains grounded in visual evidence and by training under a rubric-based reinforcement objective that scores intermediate visual and logical steps.

\subsection{Scalable Data Synthesis for Multimodal Training}
Synthetic data generation has become a standard response to annotation bottlenecks in vision-language research. LLM-driven approaches such as LLaVA-Instruct~\cite{liu2023llava}, ShareGPT4V~\cite{chen2024sharegpt4v}, and the All-Seeing project~\cite{wang2024allseeing} bootstrap instruction-following data from existing image-caption pairs, while ALLaVA~\cite{chen2024allava} further scales this strategy with GPT-4V supervision. These pipelines remain anchored to the source images they start from, limiting the diversity and visual complexity of the resulting data. Recent advances in instruction-guided image generation~\cite{podell2023sdxl, blackforestlabs2024flux, openai2025gptimage, yang2026focus} make it feasible to synthesize entirely novel scenes with controllable attributes, opening a fundamentally different scaling regime. Genesis builds on this opportunity through two complementary mechanisms. Logical answer expansion and scene variance generation turn a single seed sample into a family of training examples sharing one coherent reasoning chain. The resulting data is further coupled with rubric-based reinforcement learning that scores intermediate reasoning steps for process-level supervision.

%% file: chapters/3_vap.tex
\section{Visual Augmentation Primitives}
\label{sec:method:pipeline}

\begin{figure}
\centering
\includegraphics[width=\linewidth]{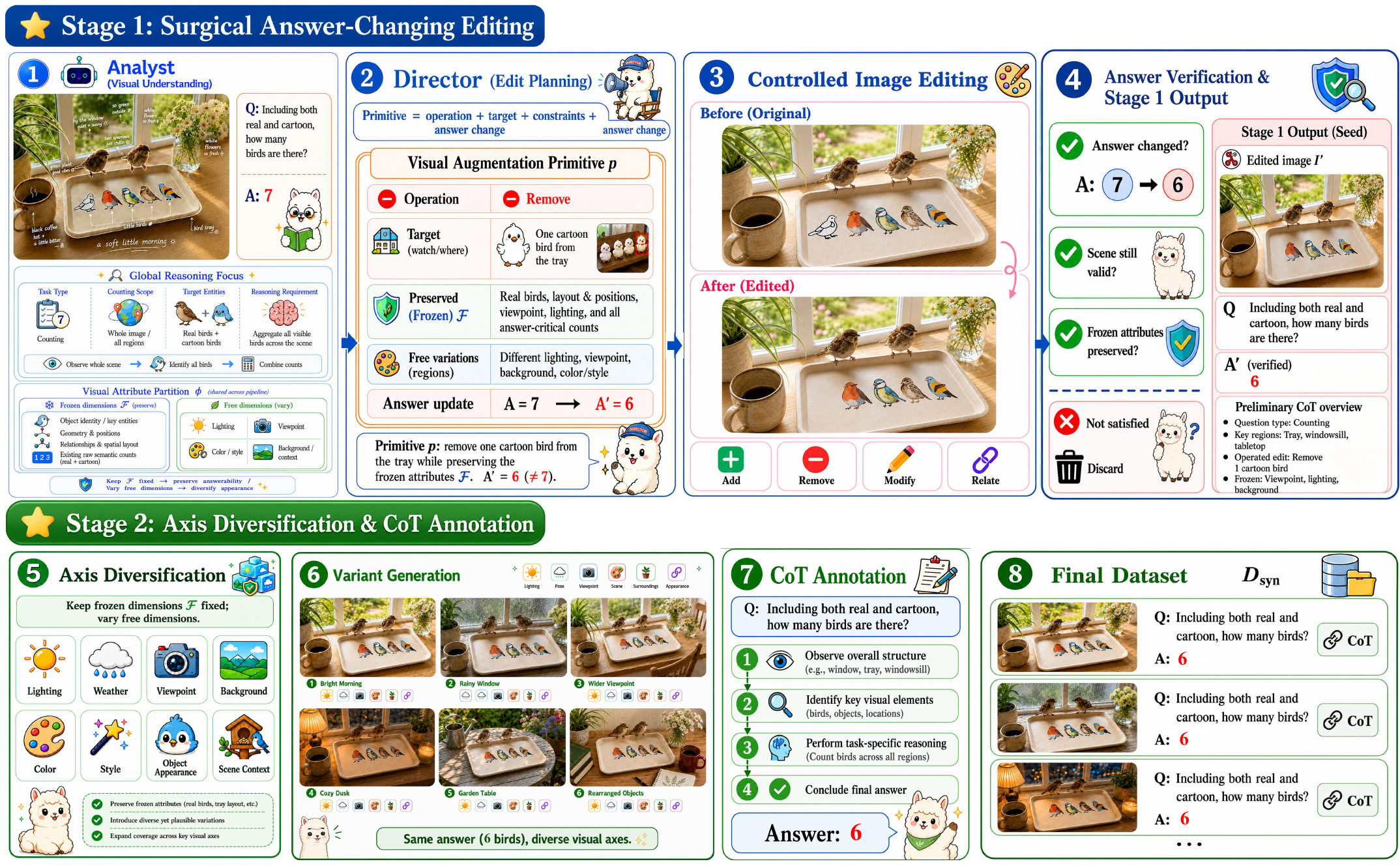}
\caption{Two-stage data construction pipeline. \textbf{Stage~1 (Surgical Answer-Changing Editing)}: the Analyst extracts the global reasoning focus and partitions visual attributes into frozen and free dimensions, the Director defines an edit primitive, the Generator executes it, and the Verifier validates the output. \textbf{Stage~2 (Axis Diversification \& CoT Annotation)}: the Planner selects free dimensions to diversify, the Generator produces visual variants, a four-step CoT is constructed for each, and the final dataset $\mathcal{D}_{\text{syn}}$ is assembled.}
\label{fig:pipeline}
\vspace{-3mm}
\end{figure}

In this chapter, we present Visual Augmentation Primitives and its two-stage data pipeline. The following chapters cover multimodal chain-of-thought rubric-based reinforcement learning (Chapter~\ref{sec:method:rl}) and the UNREAL benchmark (Chapter~\ref{sec:method:bench}), which is built by the same generative pipeline as an independent evaluation suite for hyperrealistic visual reasoning.

Current training data does not require each reasoning step to depend on visual observations, leaving complex visual reasoning capability inadequately trained.   We introduce \textbf{Visual Augmentation Primitives} (VAP), a set of controllable scene editing primitives. Instruction-guided generative models apply these primitives to modify real-world images, producing diverse photorealistic variants where each correct answer is tied to observable changes in the scene.

The VAP-based pipeline operates in two stages coordinated by five functional modules, with the \textbf{Analyst} and \textbf{Planner} instantiated by Gemini~\cite{google2025gemini35}, the \textbf{Director} by Qwen~\cite{qwen2025qwen37}, the \textbf{Generator} by GPT-Image~\cite{openai2025gptimage}, and a rule-based \textbf{Verifier} for quality control (Figure~\ref{fig:pipeline}).

\paragraph{Surgical Answer-Changing Editing (Stage 1).}
For each input triplet $(I, q, a)$, the \textbf{Analyst} examines the image to extract the global reasoning focus, including the task type, counting scope, target entities, and reasoning requirement. It then partitions the visual attributes (lighting, viewpoint, object color, background, etc.; full list in Appendix~\ref{appendix:axes}) into \textbf{frozen dimensions} ($\mathcal{F}$) that encode reasoning-critical information and must be preserved, and complementary \textbf{free dimensions} that can vary to introduce diversity. A predicate $\phi$ enforces this partition,
\begin{equation}
\phi(I') = \prod_{j \in \mathcal{F}} \mathbb{I}\left(x_j(I') = x_j(I)\right),
\label{eq:phi}
\end{equation}
where $x_j(\cdot)$ extracts the $j$-th attribute. $\phi$ serves both as a validation standard for generated images and as the basis for RL rubrics in Chapter~\ref{sec:method:rl}. The \textbf{Director} then defines a Visual Augmentation Primitive $p$ consisting of an operation from $\{\text{Add}, \text{Remove}, \text{Modify}, \text{Relate}\}$, the target element, the constraints on frozen attributes, and the corresponding answer change. The \textbf{Generator} executes the controlled edit to synthesize the edited image $I'$ with high visual fidelity while respecting $\phi$. The \textbf{Verifier} retains $(I', q, a')$ only when the answer has changed ($a' \neq a$), the edited scene remains visually plausible, and the frozen attributes are preserved ($\phi(I')=1$). This stage yields edited seed images with verified answers and a preliminary chain-of-thought overview capturing the question type, key reasoning regions, and the applied transformation.

\paragraph{Axis Diversification and CoT Annotation (Stage 2).}
Building on each verified edit from Stage~1, this stage produces visually diverse variants and constructs structured multimodal reasoning chains. The \textbf{Planner} selects which free dimensions to vary, such as lighting, weather, viewpoint, background, color, style, object appearance, and scene context, while keeping all frozen dimensions $\mathcal{F}$ unchanged. The \textbf{Generator} synthesizes multiple variants per edited sample that share the same answer but differ along the chosen free axes. For each variant, we construct a four-step multimodal chain-of-thought $z'$ that decomposes the reasoning process into (1) observing the overall image structure, (2) identifying key visual elements, (3) performing task-specific reasoning such as counting or spatial analysis, and (4) concluding with a final answer. The \textbf{Verifier} validates the quality of both the generated images and the reasoning chains, ensuring that visual variations remain faithful to $\phi$. The final dataset $\mathcal{D}_{\text{syn}}$ pairs each variant with its question, answer, and chain-of-thought.

%% file: chapters/4_rl.tex
\section{Multimodal Chain-of-Thought Rubric-Based Reinforcement Learning}
\label{sec:method:rl}

\begin{figure}[t]
    \centering
    \includegraphics[width=0.82\linewidth]{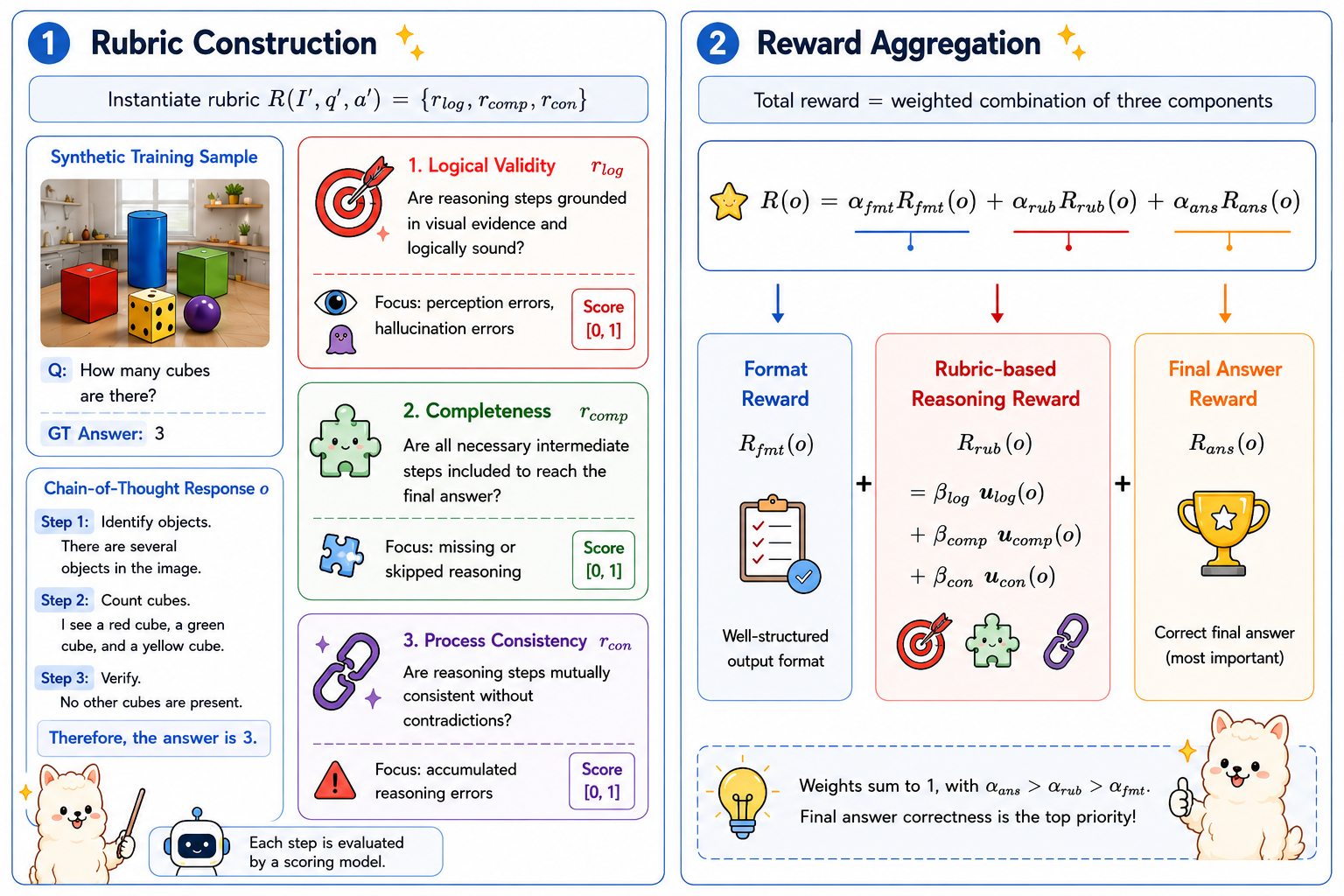}
    \caption{Overview of multimodal chain-of-thought rubric-based reinforcement learning. \textbf{(1) Rubric construction} scores each reasoning step along three dimensions (logical validity, completeness, process consistency). \textbf{(2) Reward aggregation} combines format, rubric-based reasoning, and final-answer rewards, with final-answer correctness as the top priority.}
    \label{fig:rl}
\end{figure}

Standard reinforcement learning with verifiable rewards (RLVR) optimizes only final-answer correctness, providing no reward signal on the quality of intermediate reasoning steps. The multimodal reasoning chains produced by Visual Augmentation Primitives (observation, identification, task-specific operations, and conclusion) demand process-level supervision that goes beyond binary rewards over final answers. Extending rubric-based reward design~\cite{gunjal2025rubrics, lightman2024lets, wang2024mathshepherd} from the text-only domain to visual reasoning, we introduce multimodal chain-of-thought rubric-based reinforcement learning, which evaluates each intermediate reasoning step for logical validity, completeness, and consistency, as illustrated in Figure~\ref{fig:rl}.

\paragraph{Rubric Construction.}
We score every response along three dimensions tailored to multimodal reasoning chains.
\begin{itemize}
\item \textbf{Logical validity}. Whether each step is a valid logical consequence of prior steps and visual evidence.
\item \textbf{Completeness}. Whether each step in the response aligns with the corresponding step in the ground-truth reasoning chain $z'$, providing dense step-level rewards rather than a single trajectory-level score.
\item \textbf{Process consistency}. Whether consecutive steps form a coherent chain that progresses naturally, free of contradictions or circular reasoning.
\end{itemize}
Together, these three dimensions form the rubric $\mathcal{R}(I', q', a') = \{r_{\text{log}}, r_{\text{comp}}, r_{\text{con}}\}$, with each item scored on $[0, 1]$ by an LLM-based judge.

% [benchmark figure moved to UNREAL chapter]

\paragraph{Reward Aggregation.}
The total reward combines a format reward $R_{\mathrm{fmt}}$, the rubric-based reasoning reward $R_{\mathrm{rub}}(o) = \beta_{\text{log}} u_{\text{log}}(o) + \beta_{\text{comp}} u_{\text{comp}}(o) + \beta_{\text{con}} u_{\text{con}}(o)$, and a final-answer reward $R_{\mathrm{ans}}$,
\begin{equation}
R(o) = \alpha_{\mathrm{fmt}} R_{\mathrm{fmt}}(o) + \alpha_{\mathrm{rub}} R_{\mathrm{rub}}(o) + \alpha_{\mathrm{ans}} R_{\mathrm{ans}}(o),
\label{eq:reward}
\end{equation}
with weights summing to one and ordered as $\alpha_{\mathrm{ans}} > \alpha_{\mathrm{rub}} > \alpha_{\mathrm{fmt}}$ so that final-answer correctness remains dominant. Starting from the SFT model, we optimize the policy with GRPO~\cite{shao2024deepseekmath}, using group relative advantage estimation and KL regularization toward the SFT baseline.

% \paragraph{Reward Aggregation.}
% The total reward combines a format reward $R_{\mathrm{fmt}}$, the rubric-based reasoning reward $R_{\mathrm{rub}}(o) = \beta_{\text{log}} u_{\text{log}}(o) + \beta_{\text{comp}} u_{\text{comp}}(o) + \beta_{\text{con}} u_{\text{con}}(o)$, and a final-answer reward $R_{\mathrm{ans}}$,
% \begin{equation}
% R(o) = \alpha_{\mathrm{fmt}} R_{\mathrm{fmt}}(o) + \alpha_{\mathrm{rub}} R_{\mathrm{rub}}(o) + \alpha_{\mathrm{ans}} R_{\mathrm{ans}}(o),
% \label{eq:reward}
% \end{equation}
% with weights summing to one and ordered as $\alpha_{\mathrm{ans}} > \alpha_{\mathrm{rub}} > \alpha_{\mathrm{fmt}}$ so that final-answer correctness remains dominant. Starting from the SFT model, we then optimize the policy with PPO-style updates using group relative advantage estimation and KL regularization toward the SFT baseline.

% % \paragraph*{奖励聚合。}
% % 总奖励由格式奖励 $R_{\mathrm{fmt}}$、基于评分标准的推理奖励 $R_{\mathrm{rub}}(o) = \beta_{\text{log}} u_{\text{log}}(o) + \beta_{\text{comp}} u_{\text{comp}}(o) + \beta_{\text{con}} u_{\text{con}}(o)$ 与最终答案奖励 $R_{\mathrm{ans}}$ 加权组合而成，如式~\eqref{eq:reward} 所示，权重归一化为 1 且满足 $\alpha_{\mathrm{ans}} > \alpha_{\mathrm{rub}} > \alpha_{\mathrm{fmt}}$，从而保持最终答案正确性作为主导目标。我们以 SFT 模型为起点，再以 PPO 风格更新结合组内相对优势估计和指向 SFT 基线的 KL 正则项进行优化。

%% file: chapters/5_unreal.tex
\section{UNREAL Benchmark}
\label{sec:method:bench}

\begin{figure}[t]
\centering
\includegraphics[width=0.55\linewidth]{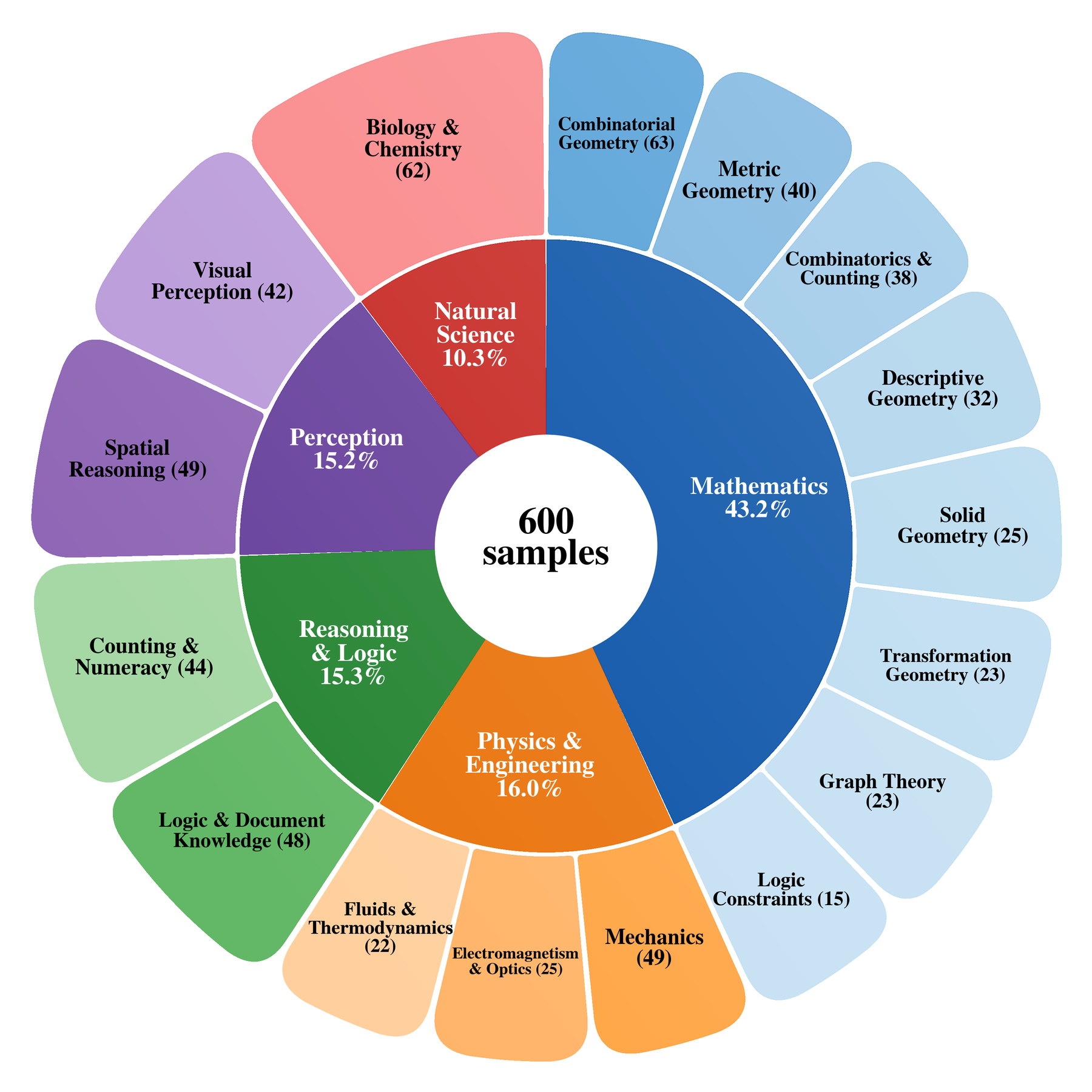}
\caption{UNREAL benchmark composition. The inner ring shows the five STEM families with their shares of the 600 samples. The outer ring shows the sixteen sub-domains with sample counts.}
\label{fig:benchmark}
\vspace{-3mm}
\end{figure}

We construct \textbf{UNREAL}, a STEM visual reasoning benchmark containing 600 samples across five STEM families (Figure~\ref{fig:benchmark}), designed to systematically evaluate model reasoning on hyperrealistic scenes. The set contains 458 open-ended questions (76.3\%) and 142 multiple-choice questions (23.7\%), enabling evaluation of both hallucination robustness~\cite{li2023pope, guan2024hallusionbench} and real-world logical reasoning. UNREAL is built around three design principles. \textbf{Rule competition} ensures the correct answer depends on the rule shown in the image rather than the model's textbook prior. \textbf{Structure preservation} keeps the logical structure of each problem fixed across visual variants. \textbf{High difficulty} requires genuine visual understanding rather than language-only pattern matching.

% \begin{figure}[t]
% \centering
% \includegraphics[width=\linewidth]{images/benchmark_v3.png}
% \caption{UNREAL benchmark composition. The inner ring shows five STEM families and the outer ring shows sub-domain distributions with sample counts.}
% \label{fig:benchmark}
% \end{figure}

\paragraph{Data Construction and Quality Control.}
UNREAL is constructed through a three-phase process. First, seed questions are collected from open educational resources across five STEM families. Mathematics (43.2\%) covers eight sub-domains including combinatorial geometry, metric geometry, and graph theory. Physics \& Engineering (16.0\%) spans mechanics, electromagnetism \& optics, and fluids \& thermodynamics. Reasoning \& Logic (15.3\%) includes logic \& document knowledge and counting \& numeracy. Perception (15.2\%) covers spatial reasoning and visual perception. Natural Science (10.3\%) addresses biology \& chemistry. Second, the VAP pipeline generates hyperrealistic scene variants for each seed with difficulty controlled along two dimensions: \textbf{reasoning complexity}, determined by the number of frozen dimensions $|\mathcal{F}|$, and \textbf{visual distraction intensity}, determined by the magnitude of variation over free dimensions. Third, every generated sample undergoes human verification for answer correctness, visual plausibility, and reasoning consistency. Samples that fail any criterion are discarded, and evaluation questions are drawn from an independent pool fully isolated from model training.

\paragraph{Evaluation Protocol.}
The protocol requires models to produce final answers directly without auxiliary cues. Given the evaluation set $\mathcal{D}_{\text{eval}} = \{(I'_n, q'_n, a'_n)\}_{n=1}^{M}$, the score of a model $\pi$ is
\begin{equation}
\mathrm{Score}(\pi) = \frac{1}{M} \sum_{n=1}^{M} \mathbb{I}\!\left( \mathrm{match}\!\left(\pi(I'_n, q'_n),\; a'_n\right) \right),
\label{eq:score}
\end{equation}
where $\mathrm{match}(\cdot,\cdot)$ denotes semantic equivalence that ignores surface differences such as numerical formatting and letter case. We report both overall and difficulty-stratified performance to analyze how model behavior scales across varying task complexity.

% To complement Genesis training with a systematic evaluation, we construct \textbf{UNREAL}, a high-difficulty STEM visual reasoning benchmark covering the four STEM families used in training (physics, biology, chemistry, and mathematics), with every question presented in open-ended form. UNREAL is built through the same generation pipeline as the training data but under strict evaluation independence, enforced through three mechanisms. Evaluation questions are drawn from an independent pool, every sample undergoes human verification, and the benchmark is fully isolated from model training.

%% file: chapters/6_experiments.tex
\section{Experiments}
\label{sec:experiments}

\newcommand{\up}[1]{$_{\textcolor{green!50!black}{\scriptstyle \mathbf{+#1}}}$}
\newcommand{\down}[1]{$_{\textcolor{red}{\scriptstyle \mathbf{-#1}}}$}

\begin{table}[t]
    \centering
    \small
    \setlength{\tabcolsep}{3.0pt}
    \caption{\textbf{Main results on STEM visual reasoning benchmarks.} \textbf{Bold} and \underline{underlined} denote the best and second-best results among open-source models, respectively. Closed-source model results are shown for reference only and are excluded from ranking. Subscripts show the improvement over the corresponding Qwen3-VL baseline. ``--'' indicates a result that is still being verified.}
    \label{tab:stem}
    \begin{tabular}{l cccccccc}
    \toprule
    Model & ZeroBench & ZB-Sub & MathVista & MMMU & MMMU-Pro & EMMA & LogicVista & BabyVision \\
    \midrule
    \multicolumn{9}{l}{\textit{Closed-source models}} \\
    GPT-5               & 2 & - & 81.3 & 84.2 & 78.4 & - & 71.8 & - \\
    Gemini-2.5-Pro      & 3 & - & 82.7 & 81.7 & 68.8 & - & 72.0 & - \\
    Claude-Opus-4.1     & 3 & - & 75.5 & 78.4 & 64.8 & - & 67.3 & - \\
    \midrule
    \multicolumn{9}{l}{\textit{Open-source models}} \\
    InternVL3-8B        & - & - & 71.6 & 62.7 & - & - & 44.1 & - \\
    InternVL3-38B       & - & - & 75.1 & \underline{70.1} & - & - & 58.4 & - \\
    LLaVA-OV-72B        & - & - & 67.5 & 56.8 & - & - & - & - \\
    Qwen3-VL-4B         & 0 & 13.2 & 74.3 & 63.3 & 36.0 & 30.5 & 50.3 & 15.0 \\
    Qwen3-VL-8B         & 0 & 18.3 & 78.2 & 63.6 & 41.4 & 31.0 & 55.9 & 10.8 \\
    Qwen3-VL-32B        & \underline{1} & \underline{26.7} & \textbf{82.9} & 68.2 & 47.9 & \underline{38.1} & \underline{63.5} & \underline{16.0} \\
    \midrule
    \rowcolor{ourscolor} \textbf{Genesis-4B}  & 0 & 17.1\up{3.9} & 77.6\up{3.3} & 63.4\up{0.1} & 53.9\up{17.9} & 32.8\up{2.3} & 60.0\up{9.7} & \textbf{16.8}\up{1.8} \\
    \rowcolor{ourscolor} \textbf{Genesis-8B}  & \underline{1}\up{1} & 24.9\up{6.6} & 78.8\up{0.6} & 66.0\up{2.4} & \underline{56.7}\up{15.3} & 35.8\up{4.8} & 59.7\up{3.8} & 15.2\up{4.4} \\
    \rowcolor{ourscolor} \textbf{Genesis-32B} & \textbf{3}\up{2} & \textbf{29.9}\up{3.2} & \underline{82.4}\down{0.5} & \textbf{75.4}\up{7.2} & \textbf{66.5}\up{18.6} & \textbf{46.5}\up{8.4} & \textbf{68.5}\up{5.0} & 14.4\down{1.6} \\
    \bottomrule
    \end{tabular}
    \vspace{-3mm}
    \end{table}

\subsection{Experimental Setup}
\label{sec:exp:setup}

\paragraph{Models and Training.}
We validate our approach on three model configurations of Qwen3-VL-Instruct at 4B, 8B, and 32B, comparing each against its corresponding base model for direct, scale-matched evaluation. All models are trained on the synthetic dataset $\mathcal{D}_{\text{syn}}$ produced by Visual Augmentation Primitives with the multimodal chain-of-thought rubric-based reinforcement learning described in Chapter~\ref{sec:method:rl}, using GRPO~\cite{shao2024deepseekmath} with group relative advantage estimation and KL regularization toward the base policy. The 4B and 8B variants are trained on a single 8$\times$GPU node, while the 32B variant uses 48 GPUs across six nodes with rollout tensor parallelism. Full hyperparameters are provided in Appendix~\ref{appendix:training}.

% \begin{table}[t]
% \centering
% \small
% \setlength{\tabcolsep}{6pt}
% \begin{tabular}{l ccc}
% \toprule
% Hyperparameter & 4B & 8B & 32B \\
% \midrule
% Learning rate              & $1{\times}10^{-6}$ & $1{\times}10^{-6}$ & $1{\times}10^{-6}$ \\
% Train batch size           & 32   & 32   & 96 \\
% PPO mini-batch size        & 16   & 16   & 48 \\
% PPO micro-batch / GPU      & 4    & 4    & 1  \\
% Rollout group size $n$     & 8    & 8    & 8  \\
% Max prompt length          & 4096 & 8192 & 4096 \\
% Max response length        & 8192 & 8192 & 8192 \\
% Max tokens / GPU           & 16{,}384 & 16{,}384 & 32{,}768 \\
% GPUs                       & 8 (1 node) & 8 (1 node) & 48 (6 nodes) \\
% Rollout TP                 & 1 & 1 & 2 \\
% \bottomrule
% \end{tabular}
% \caption{Training hyperparameters at three model scales. The 4B and 8B variants share the same base configuration; the 32B variant overrides batch sizes, per-GPU token budget, GPU count, and rollout tensor parallelism for multi-node training.}
% \label{tab:training_config}
% \end{table}

\subsection{STEM Benchmarks}
\label{sec:exp:stem}

We evaluate on eight challenging STEM benchmarks spanning mathematics, science, and visual logic, namely ZeroBench~\cite{roberts2025zerobench}, ZeroBench-Sub~\cite{roberts2025zerobench}, MathVista (MINI split)~\cite{lu2024mathvista}, MMMU (DEV-VAL split)~\cite{yue2024mmmu}, MMMU-Pro, EMMA~\cite{hao2025can}, LogicVista~\cite{xiao2024logicvista}, and BabyVision~\cite{chen2026babyvision}. Table~\ref{tab:stem} reports the main STEM results. Genesis delivers consistent improvements over the Qwen3-VL baselines at every model scale. Genesis-4B surpasses not only Qwen3-VL-8B but also InternVL3-8B on LogicVista ($60.0$ vs.\ $44.1$), confirming that process-level supervision enables a smaller model to outperform larger baselines. At the 32B scale, Genesis-32B achieves the best results on six of eight benchmarks among open-source models, with the largest improvements over Qwen3-VL-32B on MMMU-Pro ($+18.6$), EMMA ($+8.4$), and MMMU ($+7.2$). The improvements are most pronounced on benchmarks that require multi-step deduction (MMMU-Pro, EMMA, LogicVista), validating the design of rubric-based rewards over intermediate multimodal reasoning.

% Table~\ref{tab:stem} reports the main STEM results. Our method delivers consistent improvements over the Qwen3-VL baselines at every model scale. Our 4B model surpasses not only Qwen3-VL-8B but also InternVL2.5-8B on MMMU-Pro ($53.87$ vs.\ $38.52$) and LogicVista ($59.96$ vs.\ $53.60$), confirming that process-level supervision enables a smaller model to outperform larger baselines. At the 32B scale, our model achieves the best results on seven of eight benchmarks, surpassing GPT-4o by $+13.6$ on MMMU-Pro, $+6.2$ on EMMA, and $+5.1$ on MMMU. The improvements are most pronounced on benchmarks that require multi-step deduction (MMMU-Pro, EMMA, LogicVista), validating the design of rubric-based rewards over intermediate multimodal reasoning.

\subsection{VQA Benchmarks}
\label{sec:exp:vqa}

We further evaluate on six visual question answering benchmarks covering hallucination robustness (HallusionBench~\cite{guan2024hallusionbench}), aggregate visual skills (MMStar), and infographic understanding (InfoVQA).

\begin{table}[t]
    \centering
    \footnotesize
    \setlength{\tabcolsep}{4pt}
    \caption{\textbf{Main results on VQA benchmarks.} \textbf{Bold} and \underline{underlined} denote the best and second-best results among open-source models, respectively. Closed-source model results are shown for reference only and excluded from ranking. Subscripts show the improvement over the corresponding Qwen3-VL baseline.}
    \label{tab:vqa}
    \begin{tabular}{l ccc}
    \toprule
    Model & HallusionBench & MMStar & InfoVQA \\
    \midrule
    \multicolumn{4}{l}{\textit{Closed-source models}} \\
    GPT-5               & 65.7 & 76.4 & 79.0 \\
    Gemini-2.5-Pro      & 63.7 & 77.5 & 84.2 \\
    Claude-Opus-4.1     & 60.4 & 72.1 & 69.4 \\
    \midrule
    \multicolumn{4}{l}{\textit{Open-source models}} \\
    InternVL3-8B        & 49.9 & 68.2 & -    \\
    InternVL3-28B       & 57.1 & 71.5 & -    \\
    LLaVA-OV-72B        & -    & 66.1 & 74.9 \\
    Qwen3-VL-4B         & 71.2 & 68.5 & 79.7 \\
    Qwen3-VL-8B         & 72.2 & 72.0 & 82.0 \\
    Qwen3-VL-32B        & \textbf{75.0} & \underline{74.7} & \underline{87.4} \\
    \midrule
    \rowcolor{ourscolor} \textbf{Genesis-4B}  & 72.5\up{1.3} & 70.4\up{1.9} & 82.9\up{3.2} \\
    \rowcolor{ourscolor} \textbf{Genesis-8B}  & 73.0\up{0.8} & 72.0\up{0.0} & 84.7\up{2.7} \\
    \rowcolor{ourscolor} \textbf{Genesis-32B} & \underline{74.6}\down{0.4} & \textbf{75.9}\up{1.2} & \textbf{87.5}\up{0.1} \\
    \bottomrule
    \end{tabular}
    \vspace{-3mm}
    \end{table}
    
As shown in Table~\ref{tab:vqa}, Genesis-32B achieves the best results among open-source models on MMStar ($75.9$) and InfoVQA ($87.5$). It surpasses all closed-source baselines on HallusionBench ($74.6$ vs.\ $65.7$ for GPT-5) and InfoVQA ($87.5$ vs.\ $84.2$ for Gemini-2.5-Pro), demonstrating that rubric-based supervision improves both hallucination robustness and fine-grained visual understanding.

\begin{table}[t]
\centering
\small
\setlength{\tabcolsep}{4pt}
\caption{\textbf{Results on the UNREAL benchmark, stratified by domain.} \textbf{Bold} denotes the best result per column. Scores are accuracy (\%).}
\label{tab:unreal}
\begin{tabular}{l cccccc}
\toprule
Model & Overall & Math. & Nat.\ Sci. & Percep. & Phys.\ \& Eng. & Reas.\ \& Log. \\
\midrule
\multicolumn{7}{l}{\textit{Closed-source models}} \\
Claude-Opus-4.7     & 32.9 & 28.6 & 46.8          & 28.3 & 32.4 & 41.2 \\
Claude-Sonnet-4.6   & 30.7 & 23.4 & 45.2          & 32.6 & 35.3 & 36.1 \\
Doubao-Seed-1.8     & 35.9 & 27.5 & 45.2          & 39.1 & 44.1 & 42.3 \\
Gemini-3.1-Pro      & 37.1 & 23.2 & \textbf{51.6} & 43.5 & 53.9 & 44.3 \\
GPT-5.5             & \textbf{59.6} & \textbf{62.9} & \textbf{51.6} & \textbf{54.3} & \textbf{56.9} & \textbf{62.9} \\
Qwen3.6-Plus        & 42.1 & 40.7 & 37.7          & 32.6 & 44.1 & 55.7 \\
\midrule
\multicolumn{7}{l}{\textit{Open-source models}} \\
InternVL3.5-38B     & 18.3 & 6.8  & 43.5          & 19.6 & 28.4 & 23.7 \\
InternVL3.5-8B      & 16.6 & 9.6  & 38.7          & 14.1 & 20.6 & 20.6 \\
Qwen3-VL-32B        & 24.3 & 17.9 & 35.5          & 27.2 & 24.5 & 33.0 \\
Qwen3-VL-8B         & 17.5 & 7.9  & 38.7          & 18.5 & 26.5 & 21.6 \\
Qwen3-VL-4B         & 12.8 & 5.0  & 30.6          & 13.0 & 13.7 & 22.7 \\
Qwen3.5-27B         & 22.1 & 9.3  & 41.9          & 37.0 & 19.6 & 35.1 \\
Qwen3.5-9B          & 18.3 & 5.4  & 46.8          & 22.8 & 17.6 & 34.0 \\
Qwen3.5-4B          & 16.6 & 6.4  & 30.6          & 22.8 & 22.5 & 24.7 \\
\midrule
\rowcolor{ourscolor} \textbf{Genesis-4B}  & 14.2\up{1.4} & 6.1\up{1.1}  & 32.3\up{1.7} & 14.1\up{1.1} & 15.7\up{2.0} & 24.7\up{2.0} \\
\rowcolor{ourscolor} \textbf{Genesis-8B}  & 19.1\up{1.6} & 8.9\up{1.0}  & 40.3\up{1.6} & 20.7\up{2.2} & 28.4\up{1.9} & 23.7\up{2.1} \\
\rowcolor{ourscolor} \textbf{Genesis-32B} & 25.7\up{1.4} & 18.9\up{1.0} & 37.1\up{1.6} & 29.4\up{2.2} & 26.5\up{2.0} & 34.0\up{1.0} \\
\bottomrule
\end{tabular}
\vspace{-1mm}
\end{table}

\begin{table}[t]
\centering
\small
\setlength{\tabcolsep}{2pt}
\caption{Rubric-based RL vs.\ RLVR at 32B, holding data and optimizer constant.}
\label{tab:ablation:reward}
\begin{tabular}{l ccccccc}
\toprule
Reward (32B) & ZeroBench & ZB-Sub & MathVista & MMMU & MMMU-Pro & EMMA & LogicVista \\
\midrule
RLVR                                 & \textbf{3} & 28.44 & 82.20 & 75.22 & 66.42 & 45.23 & 66.22 \\
\rowcolor{ourscolor} \textbf{Rubric} & \textbf{3} & \textbf{29.94} & \textbf{82.40} & \textbf{75.44} & \textbf{66.47} & \textbf{46.48} & \textbf{68.46} \\
\bottomrule
\end{tabular}
\vspace{-1mm}
\end{table}

\begin{table}[!t]
\centering
\small
\setlength{\tabcolsep}{2pt}
\caption{Comparison of policy-optimization algorithms under rubric-based reward (4B).}
\label{tab:ablation:algo}
\begin{tabular}{l ccccccc}
\toprule
RL Algorithm (4B) & ZeroBench & ZB-Sub & MathVista & MMMU & MMMU-Pro & EMMA & LogicVista \\
\midrule
SAPO  & \textbf{0} & \textbf{20.96} & 77.2 & 63.39 & 52.20 & \textbf{34.61} & 59.06 \\
\rowcolor{ourscolor} \textbf{GRPO} & \textbf{0} & 17.07 & \textbf{77.6} & \textbf{63.44} & \textbf{53.87} & 32.78 & \textbf{59.96} \\
\bottomrule
\vspace{-6mm}
\end{tabular}
\vspace{-1mm}
\end{table}

\subsection{UNREAL Benchmark}
\label{sec:exp:unreal}

We evaluate on UNREAL, our STEM visual reasoning benchmark with controllable difficulty levels. Results are reported per STEM family and overall.

As shown in Table~\ref{tab:unreal}, UNREAL presents a significantly harder challenge than standard benchmarks. Among proprietary models, GPT-5.5 achieves the highest overall accuracy ($59.6$). Among open-source models, Qwen3.6-Plus leads at $42.1$ overall, while Qwen3-VL-32B achieves $24.3$. Performance is consistently lowest on the mathematics domain across all models, identifying multi-step symbolic-visual reasoning as the remaining frontier.

\subsection{Ablation Studies}
\label{sec:exp:ablation}

\paragraph{Rubric-Based RL vs. RLVR.}
Table~\ref{tab:ablation:reward} compares rubric-based RL against standard RLVR at the 32B scale, holding training data, optimizer, and compute budget constant. Rubric supervision yields consistent gains across all benchmarks, with the largest improvements on those that demand intermediate reasoning quality (LogicVista $+2.24$, EMMA $+1.25$, ZeroBench-Sub $+1.50$).

\paragraph{Choice of RL Algorithm.}
We compare GRPO and SAPO under the same rubric reward at the 4B scale. As shown in Table~\ref{tab:ablation:algo}, the two
  algorithms exhibit complementary strengths. Neither algorithm dominates across all benchmarks, indicating that the choice between GRPO and SAPO may depend on the target task distribution rather than one being universally superior under rubric-based supervision.  

\vspace{-3mm}

\paragraph{Training Stage Composition.}
We compare three training recipes on the same Qwen3-VL-32B initialization. We find that RL only achieves the strongest results, while inserting an SFT stage before RL hurts performance on math-heavy benchmarks. We attribute this to SFT pushing the policy toward imitating annotated reasoning chains too rigidly, constraining the exploration that rubric-based RL relies on for discovering alternative valid derivations. Skipping SFT preserves a broader prior over reasoning paths and allows the rubric reward to freely shape the learned strategies.

%% file: chapters/7_conclusion.tex
\section{Conclusion}
\label{sec:conclusion}
\vspace{-3mm}

In this paper, we propose Visual Augmentation Primitives, a generation-based data construction framework that synthesizes diverse photorealistic training samples with coherent multimodal reasoning chains, unifying visual perception with logical reasoning at every step. We further introduce multimodal chain-of-thought rubric-based reinforcement learning, which extends rubric-based reward design from the text-only domain to visual reasoning by scoring intermediate steps for logical validity, completeness, and consistency. To systematically evaluate this frontier, we construct the UNREAL benchmark covering five STEM families with controllable difficulty levels, enabling fine-grained analysis of both hallucination robustness and logical reasoning. Experiments show that a 4B model built on this framework surpasses 8B baselines and scales consistently to 32B across visual reasoning and VQA benchmarks.

\section*{Limitations}
\addcontentsline{toc}{section}{Limitations}
Our pipeline relies on commercial models (GPT-Image, Gemini, Qwen), limiting full reproducibility. UNREAL contains 600 samples across five STEM families, and scaling to more domains remains future work. The LLM-based rubric judge may introduce evaluation noise. We validate only on Qwen3-VL (4B, 8B, 32B), and the pipeline currently handles single-image editing without video or multi-image support.

%% file: chapters/acknowledgments.tex
\section*{Acknowledgments}
\addcontentsline{toc}{section}{Acknowledgments}
This work was supported by the National Key Research and Development Program of China (No.~2024YFE0210600), and the Natural Science Foundation of China under Grant Nos. 72225011 and 72434005.

%% file: chapters/appendix.tex
\begin{appendices}

\section{Data Sources and Construction Details}
\label{appendix:data}

\begin{table}[t]
\centering
\small
\begin{tabular}{lp{13cm}}
\toprule
\textbf{Family} & \textbf{Sub-domains} \\
\midrule
Physics (15) & Mechanics, Thermodynamics, Optics, Electromagnetism, Wave Motion, Fluid Dynamics, Kinematics, Gravitation, Nuclear Physics, Quantum Mechanics, Acoustics, Solid-State Physics, Astrophysics, Relativity, Circuit Analysis \\
\midrule
Biology (8) & Cell Biology, Genetics, Ecology, Anatomy, Microbiology, Botany, Zoology, Evolutionary Biology \\
\midrule
Chemistry (13) & Organic Chemistry, Inorganic Chemistry, Physical Chemistry, Analytical Chemistry, Biochemistry, Electrochemistry, Thermochemistry, Polymer Chemistry, Nuclear Chemistry, Environmental Chemistry, Crystallography, Spectroscopy, Chemical Kinetics \\
\midrule
Mathematics (14) & Algebra, Geometry, Calculus, Statistics, Probability, Number Theory, Trigonometry, Linear Algebra, Combinatorics, Graph Theory, Topology, Differential Equations, Set Theory, Logic \\
\bottomrule
\end{tabular}
\caption{STEM families and their sub-domains covered in our dataset.}
\label{tab:stem_domains}
\end{table}

% Across all sources our pipeline normalises questions into eight types: multiple-choice, numerical computation, symbolic reasoning, diagram interpretation, graph reading, spatial reasoning, process analysis, and comparative judgment.

\begin{figure}[!htbp]
\centering
\includegraphics[width=\linewidth]{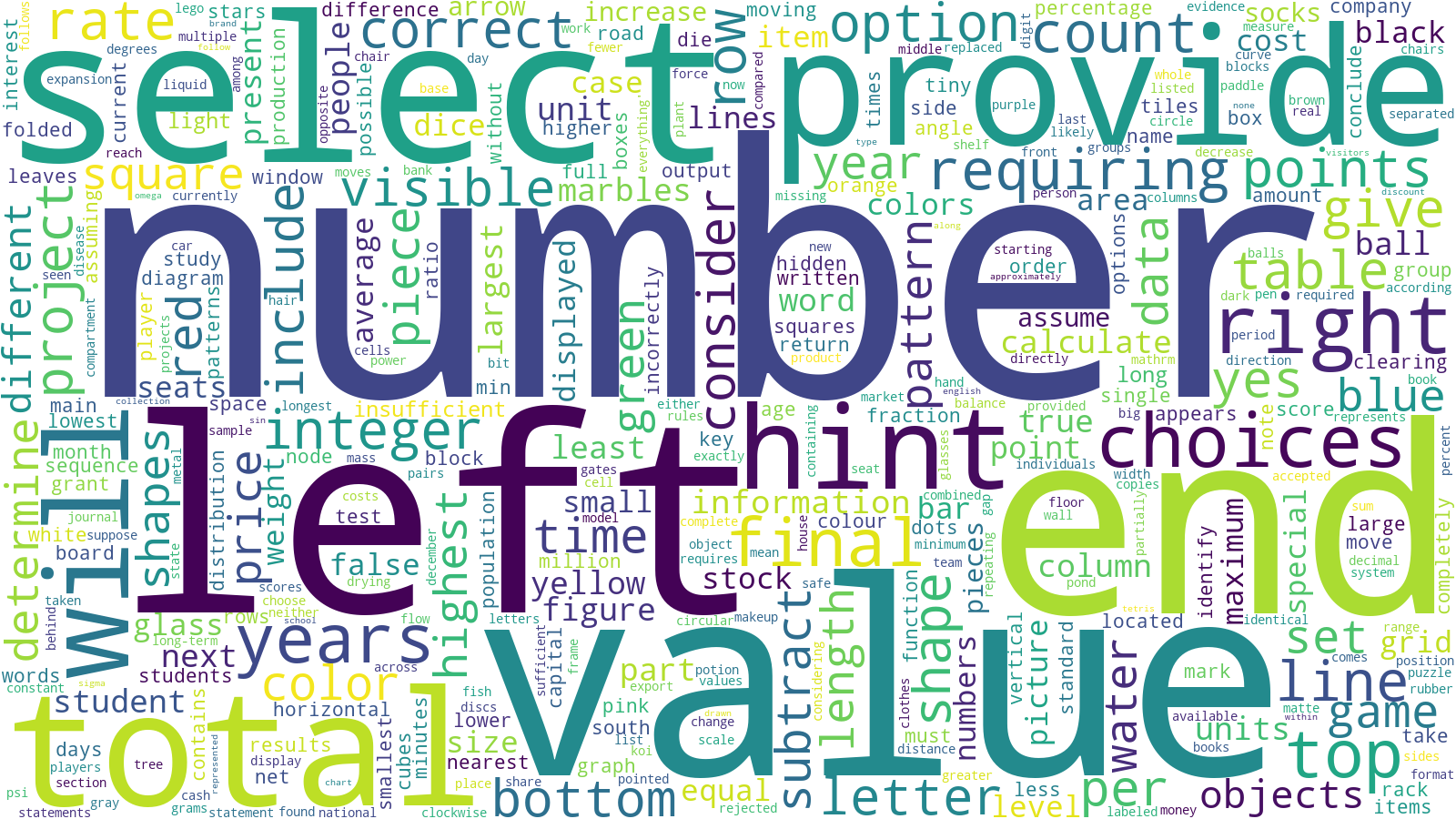}
\caption{Word cloud of question keywords in the training corpus. Word size is proportional to frequency. The distribution highlights the emphasis on quantitative reasoning (``number'', ``value'', ``total'', ``calculate''), spatial understanding (``left'', ``right'', ``top'', ``bottom''), and visual attributes (``color'', ``shape'', ``pattern'').}
\label{fig:wordcloud}
\end{figure}

Across all sources our pipeline normalises questions into eight types: multiple-choice, numerical computation, symbolic reasoning, diagram interpretation, graph reading, spatial reasoning, process analysis, and comparative judgment. Figure~\ref{fig:wordcloud} visualizes the frequency distribution of question keywords across the full training corpus, where word size is proportional to occurrence count. High-frequency terms such as ``number'', ``value'', ``total'', ``left'', and ``select'' reflect the quantitative and spatial reasoning focus of the dataset, while domain-specific terms such as ``square'', ``color'', ``diagram'', and ``pattern'' illustrate the visual grounding required by the questions.

\subsection{Visual Taxonomy}
\label{appendix:axes}

Every image is decomposed along a universal taxonomy of 19 semantic axes (the \texttt{axis\_table} produced by Call~1 of the pipeline). For exposition we group them into the six families summarised in Table~\ref{tab:visual_axes}.

\begin{table}[t]
\centering
\small
\setlength\tabcolsep{1.7pt}
\begin{tabular}{ll}
\toprule
\textbf{Category} & \textbf{Dimensions} \\
\midrule
Illumination & Lighting direction, intensity, color temperature \\
Viewpoint & Camera angle, distance, orientation \\
Object & Color, texture, shape, size, material \\
Background & Scene type, clutter level, color scheme \\
Layout & Object position, spatial arrangement, scale \\
Style & Rendering style, contrast, saturation \\
\bottomrule
\end{tabular}
\caption{Visual taxonomy used for attribute decomposition. For each image, axes are partitioned into a frozen subset $\mathcal{F}$ (must be preserved pixel-wise to keep the test point intact) and a free subset $\mathcal{V}=\mathcal{A}\setminus\mathcal{F}$ that Stage~2 aggressively varies.}
\label{tab:visual_axes}
\end{table}

For each image the pipeline classifies every axis $\alpha\in\mathcal{A}$ as \texttt{free}, \texttt{frozen}, or \texttt{depends}, conditioned on the question's \emph{test point} (the structural difficulty that makes $q$ non-trivial). Let $\phi:\mathcal{I}\to\{0,1\}$ be the frozen-attribute checker, so that $\phi(I')=1$ iff all attributes in $\mathcal{F}$ are preserved between $I$ and $I'$.

\subsection{Data Construction Pipeline}
\label{appendix:construction}

The pipeline consists of two stages built from six model calls in total. We denote a source sample as $(I,q,a)$, the Stage-1 edited image as $I'$ with rewritten answer $a'$, and the Stage-2 diversified variants as $\{I''_k\}_{k=1}^{K}$ ($K{=}4$ in our setting) which all share the same ground-truth $a'$ but differ along the free axes.

\paragraph{Stage 1: Logical Answer Expansion.}
Given an input triplet $(I,q,a)$ we proceed in five steps.
\begin{enumerate}
  \setlength{\itemsep}{0.15em}
  \item \textbf{Call~1 - Analyzer.} A Qwen-VL analyser inspects $I$ and produces an analysis $\mathcal{C}=\{\text{type}, \text{test\_point}, \mathcal{F}, \mathcal{V}, \mathcal{S}, \mathcal{E}\}$, where $\mathcal{S}$ is a set of typed slots (e.g.\ $\langle S \rangle$, $\langle SURF\rangle$, $\langle N\rangle$) and $\mathcal{E}=\{(o_i,\Delta_i,a'_i)\}$ is a candidate set of edit operations $o_i\in\{$\textsc{Remove}, \textsc{Add}, \textsc{Modify}, \textsc{Substitute}, \textsc{ContextChange}$\}$ together with their mechanically derived new answer $a'_i$. Each $\Delta_i$ belongs to one of four \emph{answer-effect types} (\textsc{Delta}, \textsc{LiteralReplace}, \textsc{FlipBoolean}, \textsc{CategoricalShift}).
  \item \textbf{Edit selection.} The pipeline draws a diverse pair of edits $(o^{(1)},o^{(2)})\subseteq\mathcal{E}$ that span at least two effect types.
  \item \textbf{Call~2a - Edit Prose.} A Qwen text model rewrites the chosen edit specification into a single 150--250-word natural-language instruction $\pi$ consumable by GPT-Image-2.
  \item \textbf{Call~3a - Image Edit.} An instruction-following image generator $\mathcal{G}_{\text{edit}}\in\{$GPT-Image-2, Gemini-3-image$\}$ produces $I'=\mathcal{G}_{\text{edit}}(I,\pi)$.
  \item \textbf{Modifier verification.} Two independent Qwen-VL verifiers $\mathcal{M}_1,\mathcal{M}_2$ read $I'$ and the rewritten question $q'$ to extract the \emph{actual} new answer:

  {\footnotesize
  \begin{equation}
    \tilde a_j = \mathcal{M}_j(I',\,q'),\quad c_j = \text{conf}_j \in \{\text{low}, \text{med}, \text{high}\}.
  \end{equation}
  }
  We accept the record iff
  \begin{equation}
    \mathrm{eq}(\tilde a_1,\tilde a_2) \;\land\; \max_j c_j \geq \text{med} \;\land\; \phi(I')=1.
    \label{eq:s1_accept}
  \end{equation}
\end{enumerate}

\paragraph{Stage 2: Visual Diversification and CoT Annotation.}
Each accepted Stage-1 record $(I',q',a',\mathcal{C})$ is then expanded in five more steps.
\begin{enumerate}
  \setlength{\itemsep}{0.15em}
  \item \textbf{Selector / axis planner.} A Qwen-VL planner produces $K$ valuations $\{\boldsymbol{v}^{(k)}\}_{k=1}^{K}$ over the free axes $\mathcal{V}$, together with slot assignments $\{\boldsymbol{s}^{(k)}\}$. We optimise for inter-plan diversity in axis space while leaving every $\alpha\in\mathcal{F}$ untouched:
  \begin{equation}
    \{\boldsymbol{v}^{(k)}\} = \arg\max_{\{v\}} \sum_{k\neq k'} d_{\mathcal{V}}(\boldsymbol{v}^{(k)},\boldsymbol{v}^{(k')}).
  \end{equation}
  \item \textbf{Template instantiation.} The question template $T_q$ and answer template $T_a$ stored in $\mathcal{C}$ are slot-filled with $\boldsymbol{s}^{(k)}$:
  \begin{equation}
    q^{(k)} = T_q(\boldsymbol{s}^{(k)}),\quad a^{(k)} = T_a(\boldsymbol{s}^{(k)}, a').
  \end{equation}
  \item \textbf{Call~2b - Edit Prose.} A Qwen text model writes a 150--260-word paragraph $\pi^{(k)}$ that paints the new background, surface, lighting, viewpoint, etc.\ implied by $\boldsymbol{v}^{(k)}$, while paraphrasing $\mathcal{F}$ into a closing integrity clause.
  \item \textbf{Image generation \& verification.} The same image backend produces $I''_k = \mathcal{G}_{\text{edit}}(I',\pi^{(k)})$. A Qwen-VL verifier then answers four yes/no checks $(A_1,\ldots,A_4)$ (CLASS, RELATION, READABILITY, NATURAL), and we accept the variant iff $\bigwedge_{i=1}^{4} A_i = \text{yes}$.
  \item \textbf{Call~4 - CoT writer.} A Gemini-3.1-pro vision-language model takes $(I''_k, q^{(k)}, a^{(k)})$ together with soft hints from $\mathcal{C}.\text{disambiguators}$ and emits a brief reasoning chain $z^{(k)}$ of 3--6 sentences, ending with the literal token \texttt{**Final Answer: $a^{(k)}$**}.
\end{enumerate}
The final supervised training tuple is $(I''_k, q^{(k)}, a^{(k)}, z^{(k)})$.

\subsection{Prompt Templates}
\label{appendix:prompts}

For reproducibility, this subsection reproduces the (system, user) prompt skeletons used at every model call in Appendix~\ref{appendix:construction}. Placeholders in \texttt{\{braces\}} are filled at runtime; the slot syntax $\langle S\rangle,\langle SURF\rangle,\langle N\rangle$ is the slot vocabulary defined in Appendix~\ref{appendix:construction}.

\paragraph{Stage 1 - Call 1: Analyzer (Qwen-VL).}

\begin{promptbox}[title=System prompt]
\small
You are an analytical assistant for an image-editing data generation pipeline. Given an image and a visual question with its ground-truth answer, your job is to produce a structured analysis JSON that downstream code uses to design controlled edits.

\medskip
\textit{Core framework: 5 edit ops $\times$ 4 answer\_effect\_types.} The five ops are \texttt{remove}, \texttt{add}, \texttt{modify\_property}, \texttt{substitute\_category}, \texttt{context\_change}. The four answer-effect types are \texttt{delta}, \texttt{literal\_replace}, \texttt{flip\_boolean}, \texttt{categorical\_shift}. Every feasible edit MUST change the answer; identity-preserving operations go to Stage~2.

\medskip
\textit{Test-point analysis.} Classify the structural difficulty into one of \{dual\_value\_disambiguation, real\_vs\_depicted, partial\_visibility, category\_filtering, lookalike\_confusion, spatial\_ordering, composition\_state, text\_in\_context, other\}. The test-point structure must remain visible after every edit.

\medskip
\textit{Stage-2 schema.} Produce a \texttt{frozen\_predicate} (answer\_function, must\_preserve, verifier\_checks, slots), a slot-templated \texttt{question\_template} and \texttt{answer\_template}, and an \texttt{axis\_table} classifying each of the 19 universal axes as \texttt{free}/\texttt{frozen}/\texttt{depends}.

Return ONLY valid JSON; no markdown fences, no commentary.
\end{promptbox}

\begin{promptbox}[title=User prompt]
\small
Original question: \texttt{\{question\}}\\
Original answer: \texttt{\{answer\}}

\medskip
Analyze this image and produce JSON with EXACTLY these fields and nothing else: \texttt{image\_type}, \texttt{scene\_description}, \texttt{question\_analysis}, \texttt{test\_point\_analysis}, \texttt{disambiguators\_to\_preserve}, \texttt{feasibility\_assessment}, \texttt{feasible\_edits} (3--5 entries spanning $\geq 2$ effect types), \texttt{frozen\_predicate}, \texttt{question\_template}, \texttt{answer\_template}, \texttt{axis\_table} (all 19 axes in fixed order).
\end{promptbox}

\paragraph{Stage 1 - Call 2a: Edit-Prose Writer (Qwen text).}

\begin{promptbox}[title=System prompt]
\small
You write image editing instructions for an AI image editor (GPT-Image-2). Your output is a SINGLE FLOWING PARAGRAPH of natural English prose, 150--250 words.

\medskip
\textit{Strict format rules.} No bullet points, no labels (no `SCENE:', `EDIT:', `KEEP:'), no section headers, no line breaks. Express constraints with subordinate clauses (``while keeping\ldots'', ``without altering\ldots'', ``so that\ldots'') and emphasise critical constraints with ``exactly'', ``only'', ``still''.

\medskip
The paragraph must (1) describe the scene including the test-point structure, (2) specify the single edit operation with precise spatial detail, (3) weave the lock list into the description, and (4) close with a one-clause integrity statement.
\end{promptbox}

\begin{promptbox}[title=User prompt]
\small
SCENE CONTEXT: \texttt{\{scene\_description\}}\\
TEST POINT: \texttt{\{core\_challenge\}} \quad (must remain visible)\\
EDIT OPERATION: type=\texttt{\{op\}}, params=\texttt{\{params\}}, expected delta=\texttt{\{expected\_visual\_delta\}}\\
MUST NOT CHANGE: \texttt{\{lock\_list\} $\cup$ \{disambiguators\}}\\
INTEGRITY CONSTRAINT: \texttt{\{edit\_must\_operate\_within\}}

\medskip
Now write the single flowing paragraph for GPT-Image-2. Output ONLY the paragraph, no other text.
\end{promptbox}

\paragraph{Stage 1 - Modifier: Dual-Validator (Qwen-VL $\times$ 2).}

\begin{promptbox}[title=System prompt]
\small
You are reading the actual answer from an edited image.
\end{promptbox}

\begin{promptbox}[title=User prompt]
\small
Background: the image shown was produced by editing an original photograph.\\
The editing instruction asked to: \texttt{\{rationale\_natural\}}\\
If the edit succeeded perfectly, the answer would be: \texttt{\{designed\_answer\}}.\\
However, AI image editing may not be perfect.

\medskip
Your task: look at THIS image and determine the ACTUAL answer to the following question, based on what you literally see.\\
Question: \texttt{\{new\_question\}}

\medskip
Output strictly in this format:\\
\texttt{ACTUAL\_ANSWER: <value>}\\
\texttt{CONFIDENCE: high | medium | low}\\
\texttt{REASONING: <one sentence>}
\end{promptbox}

\paragraph{Stage 2 - Selector: Axis Planner (Qwen-VL).}

\begin{promptbox}[title=System prompt]
\small
You are an image augmentation strategy planner. You receive an already-edited image (Stage~1 output) plus a per-image analysis (\texttt{frozen\_predicate} + \texttt{axis\_table}). Your job is to produce $N$ visually-distinct edit plans for Stage~2. Each plan picks one concrete value for every FREE axis and one slot-class for every replaceable slot.

\medskip
\textit{Hard constraints:} (1) FROZEN axes must NOT be touched; (2) \texttt{verifier\_checks} must still hold true; (3) stay strictly within the natural-photo distribution; (4) locked slots keep their default class.

\medskip
\textit{Optimization:} maximise per-axis aggressiveness (A) and inter-plan diversity in axis space (B), while keeping each plan intra-coherent (C). Also emit a 3--5-tag \texttt{diversity\_signature} per plan. Return ONLY valid JSON.
\end{promptbox}

\begin{promptbox}[title=User prompt]
\small
\texttt{scene\_description}, \texttt{test\_point}, \texttt{frozen\_predicate} (answer\_function, must\_preserve, verifier\_checks, slots), and the per-image \texttt{axis\_table} are injected. Task: produce \texttt{\{N=4\}} maximally diverse edit plans, each with \texttt{slot\_assignments}, \texttt{axis\_values}, \texttt{diversity\_signature}, and a one-sentence rationale. Output JSON only.
\end{promptbox}

\paragraph{Stage 2 - Call 2b: Diversification Prose (Qwen text).}

\begin{promptbox}[title=System prompt]
\small
You write image editing instructions for an AI image editor (GPT-Image-2). Your output is a SINGLE FLOWING PARAGRAPH of 150--260 words. Every FREE axis must be addressed with at least one concrete adjective or noun phrase.

\medskip
\textit{Strength rules.} Strictly avoid softeners (``subtle'', ``slightly'', ``a touch of'', ``gentle'', ``softly'', \ldots) and prefer amplifiers (``completely'', ``entirely'', ``dramatically'', ``fully'', ``transformed into'').

\medskip
\textit{Natural-photo rule.} The result must look like a real photograph; no illustration / cel-shaded / surreal style, no dramatic single-color stage lighting, no extreme weather.

\medskip
\textit{Frozen predicate.} Paraphrase the \texttt{must\_preserve} list into the closing of the paragraph using positive language (``the scene still shows\ldots''). Do NOT mention the question or the answer.
\end{promptbox}

\begin{promptbox}[title=User prompt]
\small
BASE SCENE: \texttt{\{scene\_description\}}\\
AXIS VALUES + SLOT CLASSES TO APPLY: \texttt{\{slot\_assignments\} $\cup$ \{axis\_values\}} (every entry must appear in the prose)\\
FROZEN PREDICATE - MUST be preserved (answer logic = \texttt{\{answer\_function\}}): \texttt{\{must\_preserve\}}

\medskip
Now write the single flowing paragraph. Open with the new background and surface plus the slot subjects; cover every free axis with a concrete adjective; close with a positive paraphrase of the \texttt{must\_preserve} constraints. Output ONLY the paragraph.
\end{promptbox}

\paragraph{Stage 2 - Verifier (Qwen-VL, 4 yes/no checks).}

\begin{promptbox}[title=System prompt]
\small
You are a strict visual verifier. You will be shown an image and 4 short yes/no questions. Answer each one with a single word (yes or no). Do not explain. Do not hedge. Use only the visual evidence in the image. If a question's premise can't be evaluated, answer ``no''.
\end{promptbox}

\begin{promptbox}[title=User prompt]
\small
Look at the image and answer each question with one word: yes or no.

\medskip
Q1 (CLASS):\quad \texttt{\{class\_check\}}\\
Q2 (RELATION):\quad \texttt{\{relation\_check\}}\\
Q3 (READABILITY):\quad \texttt{\{readability\_check\}}\\
Q4 (NATURAL):\quad ``Is this a plausible real photograph (not illustration / surreal)?''

\medskip
Output EXACTLY \texttt{A1:\,...\ A2:\,...\ A3:\,...\ A4:\,...} (one yes/no per line).
\end{promptbox}

\paragraph{Stage 2 - Call 4: CoT Writer (Gemini-3.1-pro).}

\begin{promptbox}[title=System prompt]
\small
You are a careful visual reasoning assistant. You produce a brief, accurate chain of thought for a visual question, ending with the final answer.

\medskip
\textit{Style requirements.} Write in natural, flowing English (not numbered steps, not bullet lists). Be concise (3--6 short sentences). Your reasoning must lead to the supplied final answer. You may briefly note distractors or features that distinguish the target if it makes the reasoning clearer. End with the line \texttt{**Final Answer: <answer>**}.
\end{promptbox}

\begin{promptbox}[title=User prompt]
\small
Question: \texttt{\{final\_question\}}

\medskip
Look at the image and reason briefly to the answer.

\medskip
Helpful hints (use only if natural): \texttt{\{disambiguators[:3]\}}; \texttt{\{core\_challenge\}}.

\medskip
End your response with: \texttt{**Final Answer: \{final\_answer\}**}.
\end{promptbox}

% \subsection{Qualitative Examples per Source}
% \label{appendix:examples}

% To illustrate the diversity of our pipeline outputs across source datasets, Figures~\ref{fig:ex_docvqa} show one Stage-2 sample from each of the ten source pipelines listed in Table~\ref{tab:data_sources}. Each panel contains the Stage-2 edited image, the slot-instantiated question, the Gemini-written reasoning chain $z'$, and the ground-truth answer $a'$.

\begin{figure}[t]
\centering
\includegraphics[width=\linewidth]{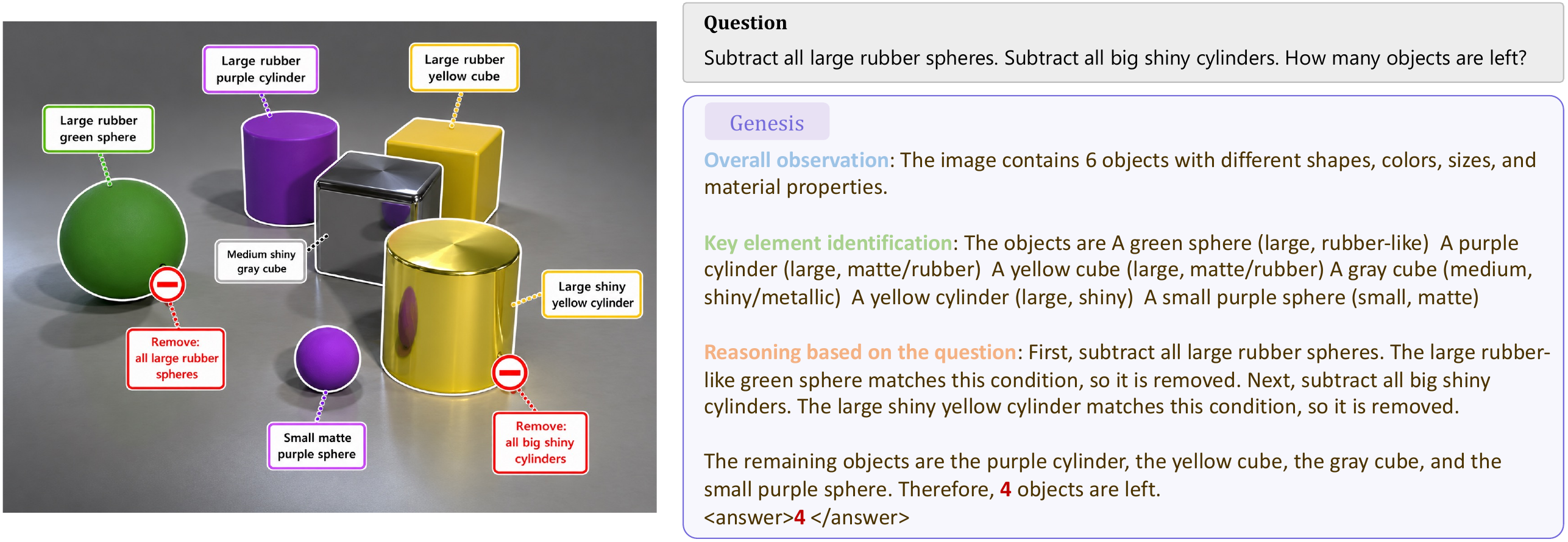}
\caption{An example of a compositional visual reasoning dataset sample, where answering the question requires reasoning over object attributes such as color, shape, size, and material.}
\label{fig:ex_docvqa}
\end{figure}

\begin{figure}[t]
\centering
\includegraphics[width=\linewidth]{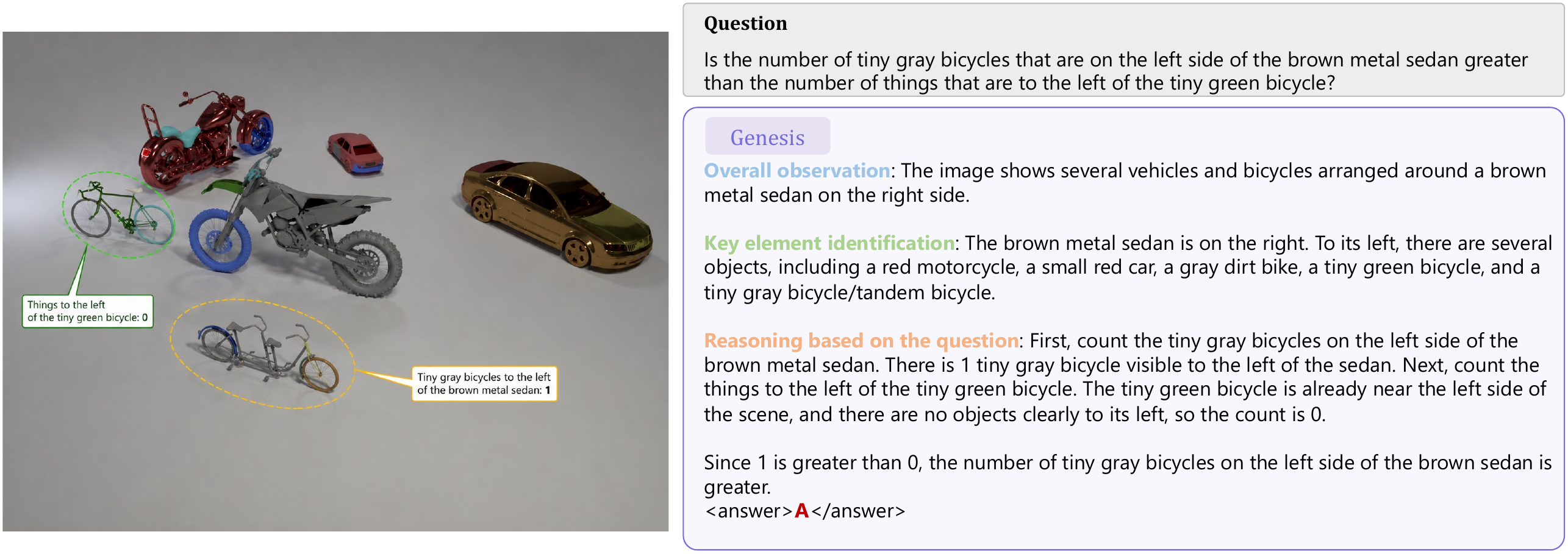}
\caption{An example of a spatial visual reasoning dataset sample, where answering the question requires reasoning over relative object positions and counting objects satisfying spatial constraints.}
\label{fig:ex_mmstar}
\end{figure}

\begin{figure}[t]
\centering
\includegraphics[width=\linewidth]{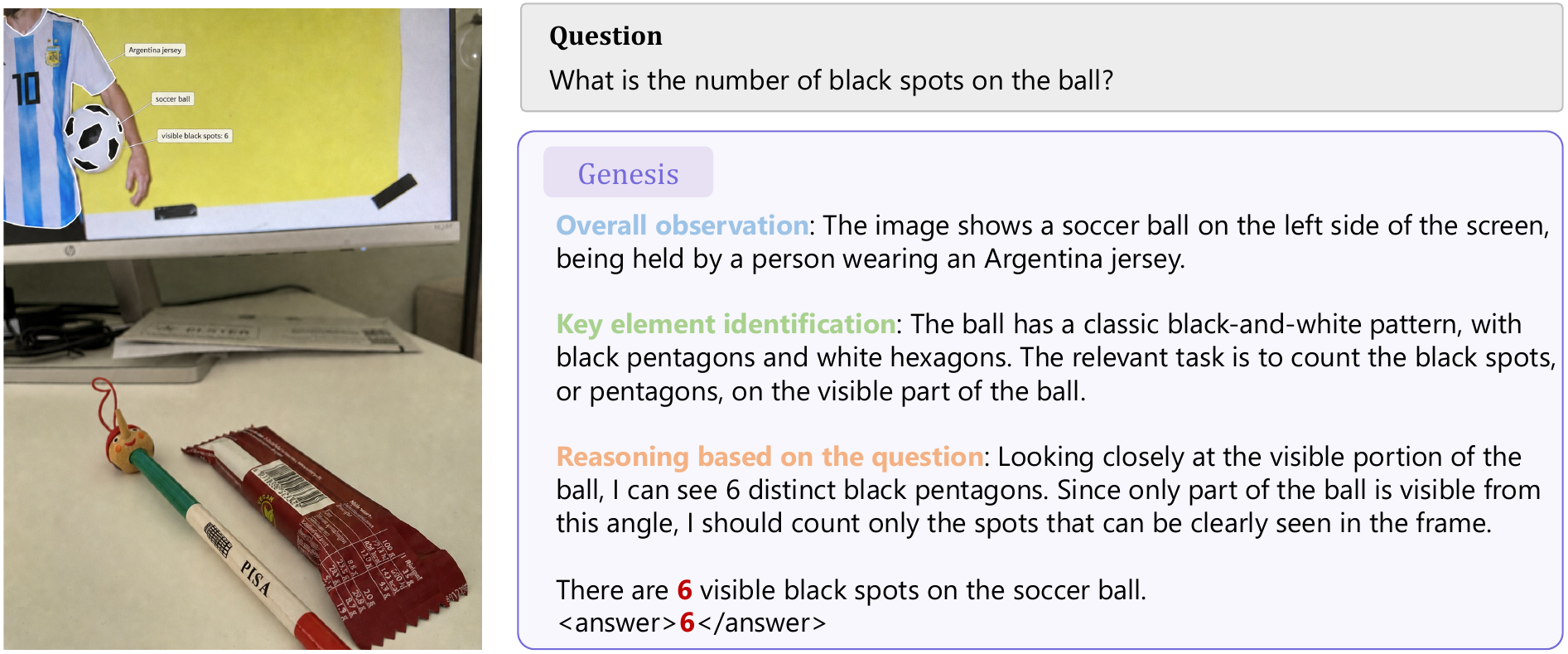}
\caption{An example of a visual counting dataset sample, where answering the question requires identifying and counting target objects in the scene.}
\label{fig:ex_infovqa}
\end{figure}

% \begin{figure}[t]
% \centering
% \includegraphics[width=\linewidth]{images/examples/example_charxiv.pdf}
% \caption{Stage-2 sample from \textbf{CharXiv} (scientific charts).}
% \label{fig:ex_charxiv}
% \end{figure}

% \begin{figure}[t]
% \centering
% \includegraphics[width=\linewidth]{images/examples/example_hallusionbench.pdf}
% \caption{Stage-2 sample from \textbf{HallusionBench} (hallucination probes).}
% \label{fig:ex_hallu}
% \end{figure}

% \begin{figure}[t]
% \centering
% \includegraphics[width=\linewidth]{images/examples/example_logicvista.pdf}
% \caption{Stage-2 sample from \textbf{LogicVista} (logical / abstract reasoning).}
% \label{fig:ex_logic}
% \end{figure}

% \begin{figure}[t]
% \centering
% \includegraphics[width=\linewidth]{images/examples/example_realworldqa.pdf}
% \caption{Stage-2 sample from \textbf{RealWorldQA} (real-world scenes).}
% \label{fig:ex_rwqa}
% \end{figure}

% \begin{figure}[t]
% \centering
% \includegraphics[width=\linewidth]{images/examples/example_emma_mini.pdf}
% \caption{Stage-2 sample from \textbf{EMMA-mini} (multimodal expert reasoning).}
% \label{fig:ex_emma}
% \end{figure}

% \begin{figure}[t]
% \centering
% \includegraphics[width=\linewidth]{images/examples/example_babyvision.pdf}
% \caption{Stage-2 sample from \textbf{BabyVision} (perceptual primitive puzzles).}
% \label{fig:ex_baby}
% \end{figure}

% \begin{figure}[t]
% \centering
% \includegraphics[width=\linewidth]{images/examples/example_in_house_stem.pdf}
% \caption{Stage-2 sample from the \textbf{In-house STEM} partition.}
% \label{fig:ex_inhouse}
% \end{figure}

\section{Multimodal Chain-of-Thought Rubric-Based Reinforcement Learning}
\label{appendix:rl}

This appendix provides the full mathematical formulation of our multimodal chain-of-thought rubric-based reinforcement learning framework. We first introduce the preliminaries of reinforcement learning with verifiable rewards (Section~\ref{appendix:rl:prelim}), then formalize rubric-based reinforcement learning (Section~\ref{appendix:rl:rubric}), and finally present our multimodal extension with detailed derivations (Section~\ref{appendix:rl:ours}).

\subsection{Preliminaries: Reinforcement Learning with Verifiable Rewards}
\label{appendix:rl:prelim}

We formulate visual question answering as a token-level Markov decision process. Given an image $I$ and a question $q$, the policy $\pi_\theta$ autoregressively generates a response $o=(o_1,o_2,\dots,o_T)\in\mathcal{V}^{\leq T_{\max}}$ over a vocabulary $\mathcal{V}$, with
\begin{equation}
\pi_\theta(o\mid I,q)=\prod_{t=1}^{T}\pi_\theta\!\left(o_t\,\middle|\,I,q,o_{<t}\right).
\label{eq:autoreg}
\end{equation}
A scalar reward $R:\mathcal{V}^{\leq T_{\max}}\to\mathbb{R}$ is assigned to the complete response.

\paragraph{Policy Optimization.}
Starting from a supervised fine-tuned (SFT) reference $\pi_{\mathrm{ref}}\!:=\!\pi_{\mathrm{sft}}$, we maximise the expected reward subject to a per-prompt KL constraint to the reference policy:

{\footnotesize
\begin{align}
\mathcal{J}(\theta)
&=\mathbb{E}_{(I,q)\sim\mathcal{D}}\,
   \mathbb{E}_{o\sim\pi_\theta(\cdot\mid I,q)}\!\left[R(o)\right] \notag\\
&\quad-\;\beta_{\mathrm{kl}}\,
   \mathbb{E}_{(I,q)\sim\mathcal{D}}\,
   D_{\mathrm{KL}}\!\bigl(\pi_\theta(\cdot\mid I,q)\,\big\|\,\pi_{\mathrm{ref}}(\cdot\mid I,q)\bigr),
\label{eq:rl_obj}
\end{align}
}
with $\beta_{\mathrm{kl}}>0$ controlling the strength of the regulariser and $\mathcal{D}$ the training distribution.

\paragraph{Group Relative Policy Optimization.}
For each prompt $(I,q)$ we sample a group $\mathcal{O}=\{o^{(i)}\}_{i=1}^{G}$ from the behaviour policy $\pi_{\mathrm{old}}$. The (response-level) advantage is normalised within the group,
\begin{equation}
\hat A^{(i)}
=\frac{R\!\left(o^{(i)}\right)-\mu_{\mathcal{O}}}{\sigma_{\mathcal{O}}+\epsilon},
\label{eq:grpo_adv}
\end{equation}
where
\(\mu_{\mathcal{O}}=\tfrac{1}{G}\sum_{i=1}^{G}R(o^{(i)})\),
\(\sigma_{\mathcal{O}}=\sqrt{\tfrac{1}{G}\sum_{i=1}^{G}\bigl(R(o^{(i)})-\mu_{\mathcal{O}}\bigr)^{2}}\),
and $\epsilon>0$ is a numerical stabiliser. The policy is updated by minimising the PPO-style clipped surrogate

{\footnotesize
\begin{align}
\mathcal{L}_{\mathrm{GRPO}}(\theta)
&=-\,\mathbb{E}_{(I,q),\,\mathcal{O}}\!\left[
   \frac{1}{G}\sum_{i=1}^{G}\frac{1}{|o^{(i)}|}
   \sum_{t=1}^{|o^{(i)}|}\!
   \min\!\Bigl(\rho^{(i)}_{t}(\theta)\,\hat A^{(i)},\right.\notag\\
&\qquad\qquad\;\;\left.\mathrm{clip}\!\bigl(\rho^{(i)}_{t}(\theta),\,1{-}\epsilon_c,\,1{+}\epsilon_c\bigr)\,\hat A^{(i)}\Bigr)
\right],
\label{eq:grpo_loss}
\end{align}
}
with per-token importance ratio
\begin{equation}
\rho^{(i)}_{t}(\theta)
=\frac{\pi_{\theta}\!\left(o^{(i)}_{t}\,\middle|\,I,q,o^{(i)}_{<t}\right)}
       {\pi_{\mathrm{old}}\!\left(o^{(i)}_{t}\,\middle|\,I,q,o^{(i)}_{<t}\right)},
\label{eq:ratio}
\end{equation}
and clip range $\epsilon_c\in(0,1)$.

\paragraph{Standard Verifiable Rewards.}
RLVR uses a binary outcome signal on the extracted final answer,
\begin{equation}
R_{\mathrm{RLVR}}(o)=\mathbf{1}\!\left[\operatorname{extract}(o)\equiv a^{\star}\right],
\label{eq:rlvr}
\end{equation}
where $\operatorname{extract}(\cdot)$ parses the answer span from $o$ and $\equiv$ denotes semantic equivalence. This signal is silent about the quality of intermediate reasoning.

\subsection{Rubric-Based Reinforcement Learning}
\label{appendix:rl:rubric}

Rubric-based reinforcement learning~\cite{gunjal2025rubrics, lightman2024lets, wang2024mathshepherd} extends RLVR with fine-grained reward signals over the quality of the reasoning process.
Specifically, a set of rubrics $\mathcal{R}=\{c_1,\dots,c_M\}$ is predefined, and a judge model $\mathcal{J}$ scores the policy model's complete output $o$ against each rubric independently:
\begin{equation}
r_m = \mathcal{J}(o \mid c_m,\, q,\, a^{\star}), \qquad m=1,\dots,M,
\label{eq:rubric_score}
\end{equation}
and the trajectory-level process reward is a weighted aggregation of the per-rubric scores:
\begin{equation}
R_{\mathrm{rub}}(o) = \sum_{m=1}^{M} \beta_m\, r_m.
\label{eq:proc_reward}
\end{equation}
Compared with the binary signal $R_{\mathrm{RLVR}}$ that only checks final-answer correctness, rubric-based rewards can distinguish between outputs that are answer-correct but reasoning-flawed and those that are reasoning-complete and well-grounded, thereby providing denser and more informative gradient signals for policy optimization.

\subsection{Multimodal Chain-of-Thought Rubric-Based RL}
\label{appendix:rl:ours}

We extend rubric-based reinforcement learning to the multimodal reasoning setting.
For a training instance $(I',q',a')$ with a reference reasoning chain $z'=(z'_1,\dots,z'_K)$,
a model response is decomposed as $o=(s_1,\dots,s_N,a)$.
We use Qwen3-32B as the judge model to score $o$ along three dimensions.

\smallskip
\noindent\textbf{(1) Logical Validity.}
The judge model examines each individual reasoning step $s_k$ and evaluates whether
the conclusion drawn at that step actually follows from the preceding premises or results,
producing an overall logical validity score $u_{\mathrm{log}}\in[0,1]$.

\smallskip
\noindent\textbf{(2) Completeness.}
This dimension measures how well $o$ covers the reference reasoning chain $z'$.
Let $\Pi(N,K)=\{\sigma:[K]\rightharpoonup[N]\mid\sigma\text{ monotone, partial, injective}\}$
be the set of monotone partial injections and $\operatorname{sim}(\cdot,\cdot)\in[0,1]$ a semantic similarity.
We first solve the optimal alignment
\begin{equation}
\sigma^{\star}=\operatorname*{arg\,max}_{\sigma\in\Pi(N,K)}\;
\sum_{j\in\mathrm{dom}(\sigma)}\operatorname{sim}\!\left(s_{\sigma(j)},\,z'_{j}\right),
\label{eq:alignment}
\end{equation}
and compute the completeness score $u_{\mathrm{comp}}\in[0,1]$ as the mean coverage of
reference steps under $\sigma^{\star}$; any reference step not matched by a response step
contributes zero.

\smallskip
\noindent\textbf{(3) Process Consistency.}
The judge model checks whether adjacent reasoning steps contain logical contradictions
or circular reasoning, producing an overall consistency score $u_{\mathrm{con}}\in[0,1]$.

\paragraph{Reward Aggregation.}
The three dimension scores are combined into a scalar rubric reward via weighted summation:
\begin{equation}
R_{\mathrm{rub}}(o)=\beta_{\mathrm{log}}\,u_{\mathrm{log}}
                  +\beta_{\mathrm{comp}}\,u_{\mathrm{comp}}
                  +\beta_{\mathrm{con}}\,u_{\mathrm{con}},
\label{eq:r_rub}
\end{equation}
where $\beta_{\mathrm{log}}+\beta_{\mathrm{comp}}+\beta_{\mathrm{con}}=1$ and $\beta_{\bullet}\geq 0$.
The full reward signal sums three components,

{\footnotesize
\begin{equation}
R(o)=\alpha_{\mathrm{ans}}\,R_{\mathrm{ans}}(o)
   +\alpha_{\mathrm{rub}}\,R_{\mathrm{rub}}(o)
   +\alpha_{\mathrm{fmt}}\,R_{\mathrm{fmt}}(o),
\label{eq:total_reward_appendix}
\end{equation}
}
with

{\footnotesize
\begin{equation}
R_{\mathrm{ans}}(o)=\mathbf{1}\!\left[\operatorname{extract}(o)\equiv a'\right],\qquad
R_{\mathrm{fmt}}(o)\in\{0,1\},
\label{eq:r_ans_fmt}
\end{equation}
}
and weights satisfying $\alpha_{\mathrm{ans}}>\alpha_{\mathrm{rub}}>\alpha_{\mathrm{fmt}}>0$,
so that final-answer correctness remains the dominant objective while reasoning quality
receives a meaningful gradient.

\paragraph{Training Objective.}
Starting from $\pi_{\mathrm{ref}}=\pi_{\mathrm{sft}}$ trained on the synthetic corpus $\mathcal{D}_{\mathrm{syn}}$, we minimise the GRPO loss in Eq.~\eqref{eq:grpo_loss} with the group advantages
\(\hat A^{(i)}\) of Eq.~\eqref{eq:grpo_adv} computed from the total reward $R(o^{(i)})$ in Eq.~\eqref{eq:total_reward_appendix}, and the per-prompt KL anchor of Eq.~\eqref{eq:rl_obj}:
\begin{align}
\mathcal{L}(\theta)
&=\mathcal{L}_{\mathrm{GRPO}}(\theta)
+\beta_{\mathrm{kl}}\,\mathbb{E}_{(I',q')\sim\mathcal{D}_{\mathrm{syn}}}\notag\\
&\quad\;\;\;D_{\mathrm{KL}}\!\bigl(\pi_{\theta}(\cdot\mid I',q')\,\big\|\,\pi_{\mathrm{ref}}(\cdot\mid I',q')\bigr).
\label{eq:full_obj}
\end{align}

\subsection{Training Configuration}
\label{appendix:training}

Table~\ref{tab:training_config} lists the full hyperparameters used for multimodal chain-of-thought rubric-based reinforcement learning across all model scales.

\begin{table}[t]
\centering
\footnotesize
\setlength\tabcolsep{1.7pt}
\begin{tabular}{l ccc}
\toprule
Hyperparameter & 4B & 8B & 32B \\
\midrule
Learning rate              & $1{\times}10^{-6}$ & $1{\times}10^{-6}$ & $1{\times}10^{-6}$ \\
Train batch size           & 32   & 32   & 96 \\
PPO mini-batch size        & 16   & 16   & 48 \\
PPO micro-batch / GPU      & 4    & 4    & 1  \\
Rollout group size $n$     & 8    & 8    & 8  \\
Max prompt length          & 4096 & 8192 & 4096 \\
Max response length        & 8192 & 8192 & 8192 \\
Max tokens / GPU           & 16{,}384 & 16{,}384 & 32{,}768 \\
GPUs                       & 8 (1 node) & 8 (1 node) & 48 (6 nodes) \\
Rollout TP                 & 1 & 1 & 2 \\
\bottomrule
\end{tabular}
\caption{Training hyperparameters at three model scales. The 4B and 8B variants share the same base configuration, while the 32B variant uses larger batch sizes, higher per-GPU token budgets, and multi-node training with rollout tensor parallelism.}
\label{tab:training_config}
\end{table}

% \paragraph{Comparison with Standard RLVR.}
% Replacing the binary RLVR reward in Eq.~\eqref{eq:rlvr} with the dense signal of Eq.~\eqref{eq:total_reward_appendix} yields informative gradients even when $R_{\mathrm{ans}}(o)=0$: a response with $R_{\mathrm{ans}}(o)=0$ but $R_{\mathrm{rub}}(o)>0$ still produces a positive group-relative advantage whenever its rubric score exceeds the group mean. The multimodal conditioning in Eqs.~\eqref{eq:u_log}--\eqref{eq:u_con} further ensures that step-level credit is assigned with respect to both $I'$ and $q'$, separating our approach from text-only rubric methods such as~\citet{lightman2024lets,wang2024mathshepherd}.

\section{UNREAL Benchmark Details}
\label{appendix:unreal}

\begin{figure}[t]
  \centering
  \includegraphics[width=0.72\linewidth]{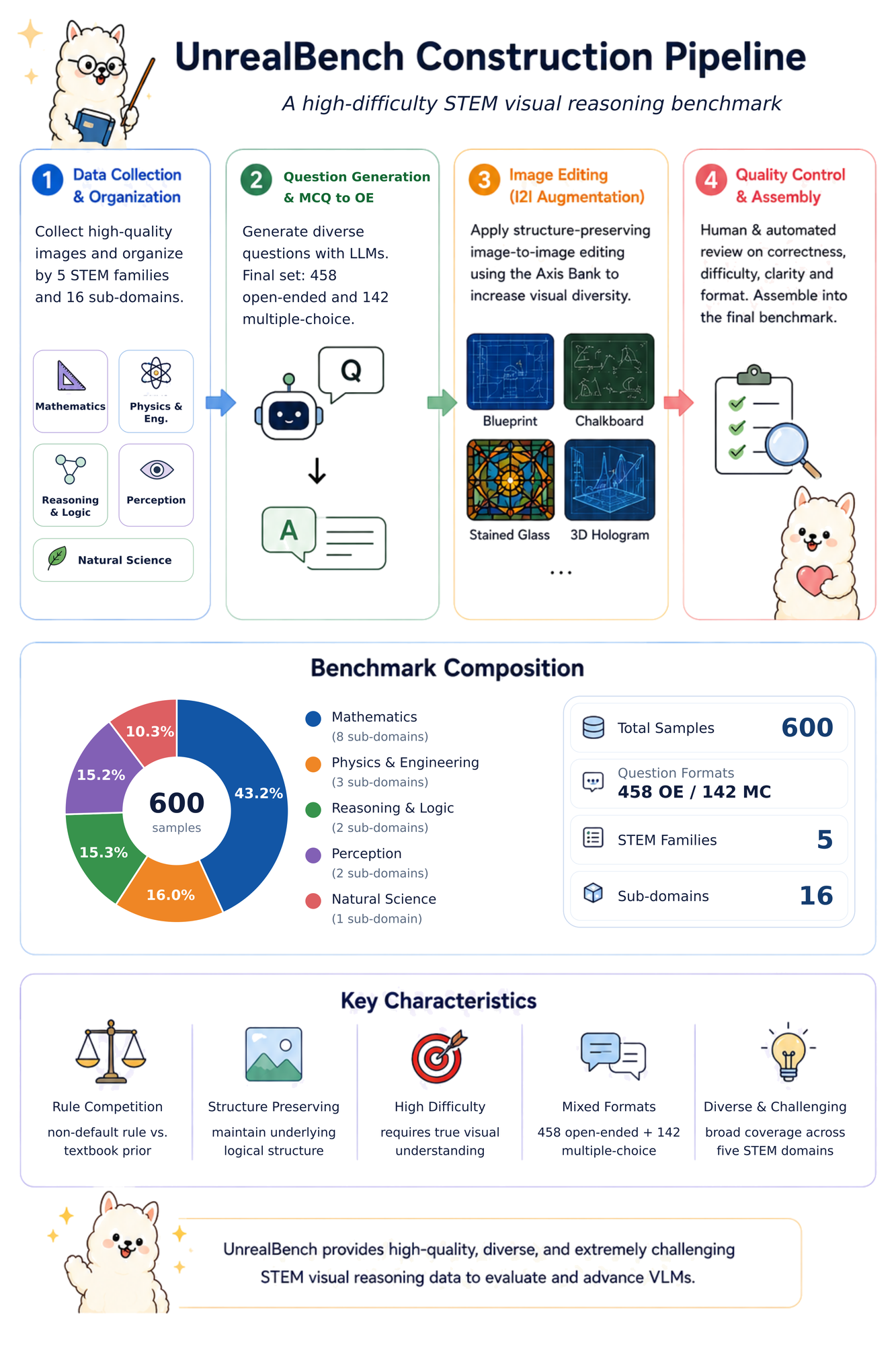}
  \caption{Overview of the UNREAL benchmark construction pipeline. \textbf{Top:} the four sequential stages, namely (1)~data        
  collection and organisation across five STEM families and sixteen sub-domains, (2)~LLM-driven question generation with MCQ$\rightarrow$OE       
  conversion, (3)~structure-preserving image-to-image augmentation using the Axis Bank, and (4)~human and automated quality       
  control. \textbf{Middle:} the benchmark composition over the five STEM families together with format and scale          
  statistics. \textbf{Bottom:} the key characteristics of the benchmark, including rule competition, structure preservation, high difficulty, mixed answer formats, and broad STEM coverage.}
  \label{fig:unreal_pipeline}
\end{figure}

This appendix provides the full topic distribution, mechanism taxonomy, and construction details of the UNREAL benchmark. A high-level summary of the entire construction pipeline, the family composition, and the key design characteristics is given in Figure~\ref{fig:unreal_pipeline}.

\subsection{Overview}
\label{appendix:unreal:overview}

\begin{figure}[t]
\centering
\includegraphics[width=0.85\linewidth]{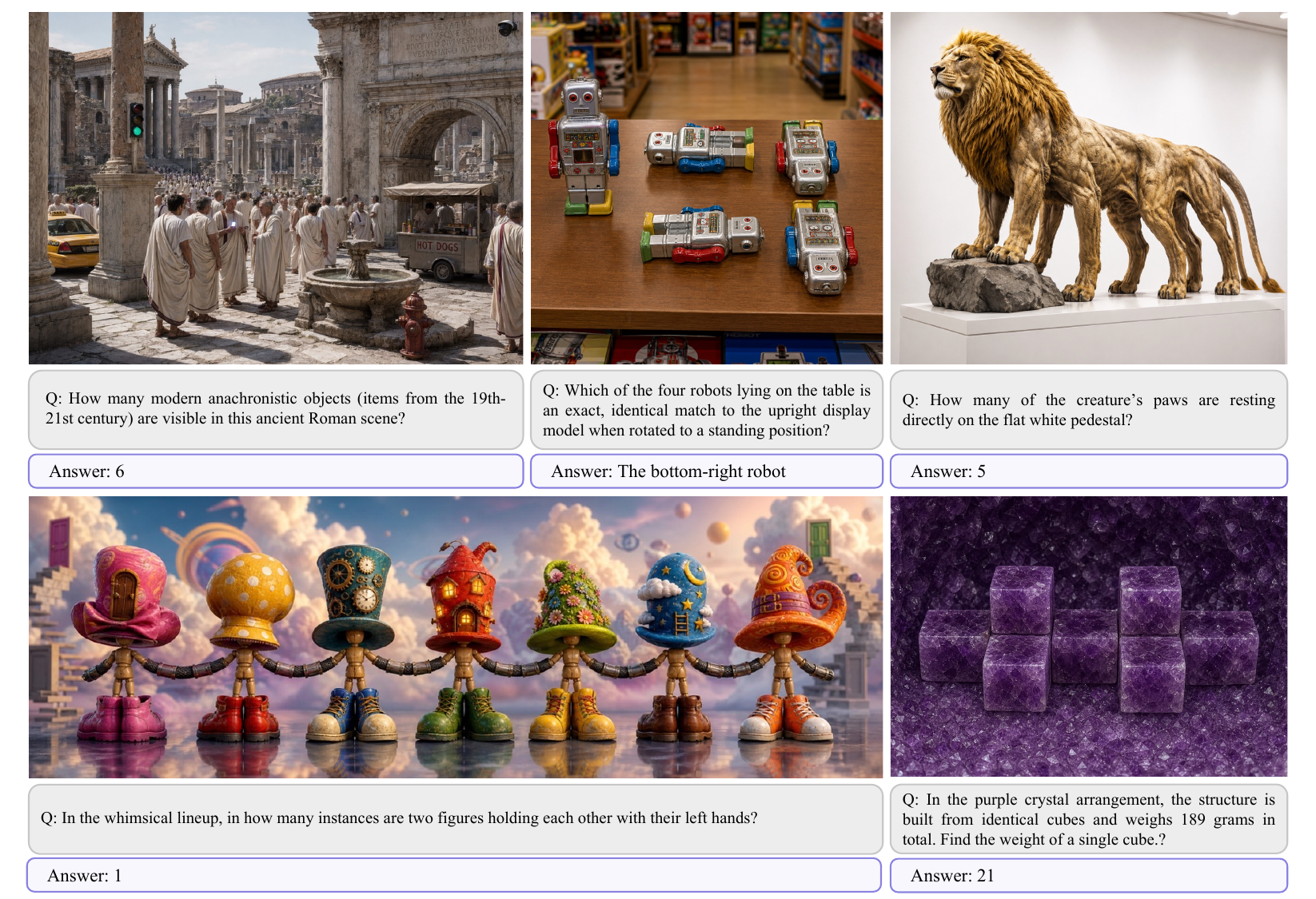}
\caption{UNREAL benchmark illustration.}
\label{fig:fig2}
\end{figure}

UNREAL contains 600 evaluation samples across five STEM families, produced via Stage~1 of the Visual Augmentation Primitives pipeline. The set combines 458 open-ended questions (76.3\%) and 142 multiple-choice questions (23.7\%). Open-ended questions require models to generate answers directly. The end-to-end pipeline is illustrated in Figure~\ref{fig:unreal_pipeline}. Figure~\ref{fig:fig2} shows some examples.

% \begin{figure}[t]
% \centering
% \includegraphics[width=0.85\linewidth]{images/unreal_overview.pdf}
% \caption{UNREAL benchmark composition across the four STEM families.}
% \label{fig:unreal_overview}
% \end{figure}

\begin{table}[t]
\centering
\small
\begin{tabular}{lcc}
\toprule
\textbf{Family} & \textbf{Samples} & \textbf{Share} \\
\midrule
Mathematics & 259 & 43.2\% \\
Physics \& Engineering & 96 & 16.0\% \\
Reasoning \& Logic & 92 & 15.3\% \\
Perception & 91 & 15.2\% \\
Natural Science & 62 & 10.3\% \\
\midrule
Total & 600 & 100\% \\
\bottomrule
\end{tabular}
\caption{UNREAL benchmark composition by STEM family.}
\label{tab:unreal_overview}
\end{table}

\begin{table}[t]
\centering
\small
\begin{tabular}{llc}
\toprule
\textbf{Family} & \textbf{Sub-domain} & \textbf{Samples} \\
\midrule
Mathematics & Combinatorial Geometry & 63 \\
 & Metric Geometry & 40 \\
 & Combinatorics \& Counting & 38 \\
 & Descriptive Geometry & 32 \\
 & Solid Geometry & 25 \\
 & Transformation Geometry & 23 \\
 & Graph Theory & 23 \\
 & Logic Constraints & 15 \\
Physics \& Engineering & Mechanics & 49 \\
 & Electromagnetism \& Optics & 25 \\
 & Fluids \& Thermodynamics & 22 \\
Reasoning \& Logic & Logic \& Document Knowledge & 48 \\
 & Counting \& Numeracy & 44 \\
Perception & Spatial Reasoning & 49 \\
 & Visual Perception & 42 \\
Natural Science & Biology \& Chemistry & 62 \\
\bottomrule
\end{tabular}
\caption{UNREAL sub-domain distribution across the five STEM families.}
\label{tab:unreal_subdomains}
\end{table}

\end{appendices}